%% file: main.tex
\documentclass[letterpaper]{article} % DO NOT CHANGE THIS
\usepackage{aaai2026}  % DO NOT CHANGE THIS
\usepackage{times}  % DO NOT CHANGE THIS
\usepackage{helvet}  % DO NOT CHANGE THIS
\usepackage{courier}  % DO NOT CHANGE THIS
\usepackage[hyphens]{url}  % DO NOT CHANGE THIS
\usepackage{graphicx} % DO NOT CHANGE THIS
\usepackage{natbib}  % DO NOT CHANGE THIS AND DO NOT ADD ANY OPTIONS TO IT
\usepackage{caption} % DO NOT CHANGE THIS AND DO NOT ADD ANY OPTIONS TO IT
\usepackage{newfloat}
\usepackage{listings}

\usepackage{newfloat}
\usepackage{listings}
\usepackage{booktabs}       % professional-quality tables
\usepackage{amsfonts}       % blackboard math symbols
\usepackage{nicefrac}       % compact symbols for 1/2, etc.
\usepackage{microtype}      % microtypography
\usepackage{xcolor}         % colors
\usepackage{array}     % 用于 \arraystretch
\usepackage{amssymb}   % 用于 \checkmark
\usepackage{graphicx}  % 可能需要，例如旋转文本
\usepackage{amssymb}   % For \checkmark (如果 \cmark, \xmark 定义依赖它)
\usepackage{pifont}   % 如果用了 \ding
\usepackage{amsmath}
\DeclareMathOperator*{\argmax}{arg\,max} % 定义一个新的 argmax 操作符
\usepackage{pifont}    % For \ding{51} (checkmark) and \ding{55} (crossmark)
\usepackage{booktabs}  % 用于三线表样式
\usepackage{threeparttable} % 用于表格注释 (如果需要解释缩写)
\usepackage{pdfpages}
\usepackage{subcaption} % 创建子图 (推荐)
\usepackage{siunitx}
\usepackage{tabularx}
\usepackage{listings}
\usepackage[table]{xcolor}
\usepackage[linesnumbered,lined,boxed,commentsnumbered,ruled]{algorithm2e}
\usepackage{lipsum} % For dummy text
\usepackage{tabularray}
\definecolor{codegreen}{rgb}{0,0.6,0}
\definecolor{codegray}{rgb}{0.5,0.5,0.5}
\definecolor{codepurple}{rgb}{0.58,0,0.82}
\definecolor{backcolour}{rgb}{0.95,0.95,0.92} % A light beige background

\lstdefinestyle{mystyle}{
    backgroundcolor=\color{backcolour},
    commentstyle=\color{codegreen},
    keywordstyle=\color{blue},
    numberstyle=\tiny\color{codegray},
    stringstyle=\color{codepurple},
    basicstyle=\ttfamily\footnotesize,
    breakatwhitespace=false,
    breaklines=true,
    captionpos=b,
    keepspaces=true,
    numbers=none,
    numbersep=5pt,
    showspaces=false,
    showstringspaces=false,
    showtabs=false,
    tabsize=4, % Adjusted tabsize
    frame=tb,  % Top and bottom frame lines
    framerule=0.5pt, % Thickness of frame rule
    rulecolor=\color{black},
    classoffset=1, % For custom keywords like classes
    morekeywords={abc, Any, Dict, List, Optional, Set, Tuple, True, False, None, self, init, str, int, float, bool}, % Add common Python types and self
    keywordstyle=\color{blue}, % Keywords color
    classoffset=0, % Reset class offset
    literate=*{_}{{\_}}1, % Treat underscore as a normal character in text mode
}
\newcommand{\cmark}{\textcolor{green!60!black}{\ding{51}}}%
\newcommand{\xmark}{\textcolor{red}{\ding{55}}}%
\newcolumntype{L}[1]{>{\raggedright\arraybackslash}p{#1}}
\newcolumntype{C}[1]{>{\centering\arraybackslash}p{#1}}

\usepackage{multirow} % 用于合并行

\title{AutoTool: Efficient Tool Selection for Large Language Model Agents}
\author{
    Jingyi Jia,
    Qinbin Li\thanks{Corresponding Author}
}
\affiliations{
    School of Computer Science and Technology\\
    Huazhong University of Science and Technology\\
    \{jingyijia, qinbin\}@hust.edu.cn
}

\usepackage{bibentry}
\begin{document}

\maketitle

\begin{abstract}
Large Language Model (LLM) agents have emerged as powerful tools for automating complex tasks by leveraging the reasoning and decision-making abilities of LLMs. However, a major bottleneck in current agent frameworks lies in the high inference cost of tool selection, especially in approaches like ReAct that repeatedly invoke the LLM to determine which tool to use at each step. In this work, we propose AutoTool, a novel graph-based framework that bypasses repeated LLM inference by exploiting a key empirical observation: tool usage inertia—the tendency of tool invocations to follow predictable sequential patterns. AutoTool constructs a directed graph from historical agent trajectories, where nodes represent tools and edges capture transition probabilities, effectively modeling the inertia in tool selection. It further integrates parameter-level information to refine tool input generation. By traversing this structured representation, AutoTool efficiently selects tools and their parameters with minimal reliance on LLM inference. Extensive experiments across diverse agent tasks demonstrate that AutoTool reduces inference costs by up to 30\% while maintaining competitive task completion rates, offering a practical and scalable enhancement for inference-heavy frameworks. Our work highlights the promise of integrating statistical structure into LLM agent design for greater efficiency without sacrificing performance. 

\begin{links}
    \link{Code}{https://github.com/jiajingyyyyyy/AutoTool}
\end{links}
% Our code is available at \url{https://github.com/jiajingyyyyyy/AutoTool}.

\end{abstract}

\section{Introduction}
Large Language Models (LLMs) have experienced explosive growth, demonstrating impressive capabilities in various tasks~\cite{achiam2023gpt, dubey2024llama, yang2025qwen3} from natural language understanding~\cite{devlin2019bert} and reasoning~\cite{guo2025deepseek} to automation of complex workflows~\cite{wang2025megaagent}. LLM-powered agents, which leverage these capabilities for interactive~\cite{yi2024survey} and decision-making~\cite{wei2025plangenllms} tasks, have become increasingly prevalent in numerous domains, including software development~\cite{jin2024llms}, intelligent personal assistants~\cite{li2024personal} and scientific research automation~\cite{team2025tongyi}.

However, despite their versatility, a significant drawback of current LLM-based agents is the substantial computational overhead~\cite{kim2025cost}, particularly evident in frameworks like ReAct~\cite{yao2023react} that involve many LLM inferences. Among these inference processes, one major goal is to repeatedly infer appropriate tools to use based on dynamic contexts, leading to high inference costs and latency. Given these challenges, a natural question arises: \emph{\textbf{Can we utilize statistical methods instead of relying heavily on LLM inference to select tools efficiently?}}

Addressing this issue, we empirically observe a critical phenomenon termed \textit{tool usage inertia}, where tool selections demonstrate predictable sequential patterns. For instance, when an agent searches in an academic database, the invocation of \textit{AuthorNodeCheck} is often followed by \textit{LoadAuthorNet} to retrieve detailed information. This sequential dependence is observable across diverse agent tasks~\cite{qin2023toolllm, ma2024agentboard}, confirming that prior tool selections significantly influence subsequent choices. Such inertia is widely present in many domain tasks, where LLM agents are usually applied.

Inspired by these insights, we introduce \textbf{AutoTool}, a novel graph-based method for automatic tool selection in LLM agents. AutoTool constructs a graph representation capturing the observed inertia in tool selection from historical workflows, where nodes correspond to tools and edges encode observed sequential dependencies. Additionally, AutoTool intelligently integrates parameter-level information into this graph, thereby enabling automated parameter filling. By efficiently traversing this structured representation, AutoTool selects the appropriate tools and parameters significantly faster than traditional inference-based approaches. Experimental evaluations demonstrate that AutoTool substantially reduces token consumption and LLM call counts while maintaining comparable agent task progress rates to LLM-based methods. 

Our contributions are summarized as follows:
\begin{itemize}
\item We empirically identify and analyze the phenomenon of tool usage inertia in LLM-based agents, both in tool selection and parameter filling.
\item We design a method to construct an inertia-aware tool graph that captures sequential patterns and data flow in agent behavior.
\item We develop a graph-based selection algorithm that efficiently determines the next tool and its parameters with minimal LLM intervention.
\item We conduct extensive experiments showing that AutoTool achieves significant reductions in LLM inference cost while preserving task performance.
\end{itemize}

% 在导言区需要 \usepackage{array}
\newcolumntype{C}[1]{>{\centering\arraybackslash}p{#1}}

\begin{table*}[t]
\centering
\renewcommand{\arraystretch}{1.2}
\setlength{\tabcolsep}{6pt}
% --- 核心修改：增加一列定义 ---
\begin{tabular}{@{} l C{1.5cm} C{1.5cm} C{1.5cm} C{1.7cm} C{1.4cm} C{1.8cm} C{2.2cm} @{}}
\toprule
\textbf{Feature} & 
\textbf{AutoTool} & 
\textbf{DFSDT} & 
\textbf{AnyTool} & 
\textbf{ToolChain}  & 
\textbf{ToolNet} & 
\textbf{ToolPlanner} &
\textbf{LLMCompiler} \\
\midrule
Efficiency       & \cmark & \xmark & \cmark & \xmark & \cmark & \cmark & \cmark \\
LLM Offloading   & \cmark & \xmark & \xmark & \xmark & \xmark & \xmark & \xmark \\
Inertia Aware    & \cmark & \xmark & \xmark & \xmark & \xmark & \xmark & \xmark \\
Parameter Flow   & \cmark & \xmark & \xmark & \xmark & \xmark & \xmark & \xmark \\
Tool Graph       & \cmark & \xmark & \cmark & \cmark & \cmark & \xmark & \cmark \\
\bottomrule
\end{tabular}
\caption{A comparison of AutoTool with other tuning-free tool selection methods}
\label{tab:sota_comparison_fixed}
\end{table*}

\section{Background and Related Work}

\subsection{LLM Agent Frameworks}

LLM agents have significant potential in solving complex problems, primarily through effective task planning, reasoning, and interaction with external tools~\cite{qin2024tool, qu2025tool}. The seminal work ReAct~\cite{yao2023react} introduces the core paradigm of driving agent decisions through interleaved ``Thought-Act-Observe" cycles, which has become a cornerstone for numerous open-source frameworks, including Langchain~\cite{langchain2023} and MetaGPT~\cite{hong2023metagpt}.
Subsequent research has expanded agents' capabilities to interact with a vast array of external tools, such as RESTful APIs in RestGPT~\cite{song2023restgpt} or various AI models orchestrated by HuggingGPT~\cite{shen2023hugginggpt}. However, a shared limitation across these powerful frameworks is that the fundamental decision of which tool to use at each step still predominantly relies on a costly LLM inference~\cite{belcak2025small}. This reliance creates a significant computational bottleneck, which is the primary issue our work aims to address.

\subsection{Automated Tool Selection}

Research in tool selection can be broadly categorized into two types: fine-tuning-dependent and tuning-free methods.
Fine-tuning methods like Toolformer~\cite{schick2023toolformer}, Gorilla~\cite{patil2024gorilla} and ToolRL~\cite{qian2025toolrl} aim to improve intrinsic tool-use capabilities. While these approaches demonstrably enhance a model's intrinsic ability to call tools correctly, their reliance on high-quality data or carefully crafted reward signals presents a significant barrier to scalability and adaptability. Tuning-free methods employ different strategies at runtime. Approaches like AnyTool~\cite{du2024anytool} and ToolNet~\cite{liu2024toolnet} focus on improving retrieval completeness and handling large-scale APIs. Many search strategies are introduced to find the optimized action sequences, such as the DFSDT in ToolLLM~\cite{qin2023toolllm}, BFS in ITS~\cite{koh2024tree}, and the toolkit-based planning in ToolPlanner~\cite{liu2024tool}.

\subsection{Automated Workflow Generation}

To accomplish many complex tasks, agents need to execute a sequence of interdependent actions. Some approaches focus on search and planning. For instance, Tree of Thoughts~\cite{yao2023tree} explores diverse action paths, while ToolChain~\cite{zhuang2023toolchain} employs the A* search algorithm over a decision tree. Other works, such as LLMCompiler~\cite{kim2024llm}, target execution efficiency by enabling parallel function calling. In parallel, another key approach is learning from past interactions: ART~\cite{paranjape2023art} automates multi-step reasoning and tool use by retrieving and composing examples from a task library, whereas Agent Workflow Memory~\cite{wang2024agent} extracts reusable workflows to guide web agents in long-horizon tasks. 
Furthermore, several works have begun to apply graph-based methods to enhance planning and workflow automation~\citep{zhuge2024gptswarm, wu2024can}. Although most of these methods primarily focus on improving success rates and workflow quality, their planning or search processes can be computationally intensive. A comparison between our method and related studies is summarized in Table~\ref{tab:sota_comparison_fixed}.

\section{Motivation}

\paragraph{Observation 1: LLM Agents have high inference costs for tool selection}
\label{sec:motivation_cost}

The predominant focus in LLM agent research has been on maximizing task success rates, often leaving significant room for improvement in operational efficiency—a crucial factor for practical deployment. A multi-step task can trigger numerous LLM calls, posing significant challenges for real-time or resource-constrained applications. 

We argue that this universal reliance on the LLM is not only costly but also \emph{sub-optimal}. The core issue is that not all decision steps in a task are of equal complexity or importance. Many tool invocations, both for selection and parameter filling, occur in highly patterned or repetitive contexts that do not require the LLM's full, nuanced reasoning power. This universal application of a resource-intensive model for both complex and simple decisions constitutes an over-utilization of its powerful abilities, creating unnecessary computational overhead.

\paragraph{Observation 2: Tool Invocation Exhibits Predictable, Low-Entropy Inertia}
\label{sec:motivation_inertia}

Motivated by this efficiency challenge, we conducted an empirical analysis to test the hypothesis that tool usage is not a series of independent events, but rather a process governed by sequential patterns. We utilize a ReAct agent within the ScienceWorld environment to generate 322 trajectories, yielding 6014 tool invocations. Analyzing these sequences, we discover strong ``Sequential Inertia": tool selection is not independent but follows predictable, low-entropy patterns.

\begin{table}[t]
\centering
\small
\renewcommand{\arraystretch}{1.1}
\setlength{\tabcolsep}{5pt}

\begin{tabular}{@{} l l l S[table-format=2.1] @{}}
\toprule
\textbf{Action} & \textbf{Src Tool} & \textbf{Src Param} & {\textbf{\%}} \\
\midrule
\multirow{5}{*}{\texttt{use(target)}} 
 & \texttt{move}       & \texttt{source} & 44.8 \\ % <-- \_
 & \texttt{pick\_up}  & \texttt{OBJ}         & 22.1 \\ % <-- \_
 & \texttt{move}       & \texttt{target}   & 11.5 \\ % <-- \_
 & \texttt{pour}      & \texttt{OBJ}    &  11.0 \\ % <-- \_
 & Other              & --                   &  10.6 \\
\midrule
\multirow{5}{*}{\texttt{pick\_up(OBJ)}}
 & \texttt{focus\_on}  & \texttt{OBJ}   & 40.1 \\ % <-- \_
 & \texttt{move}      & \texttt{source}   & 24.3 \\ % <-- \_
 & \texttt{pour}      & \texttt{from}     & 16.4 \\ % <-- \_
 & \texttt{look\_at}    & \texttt{OBJ}     & 5.9 \\
 & Other              & --                   & 13.3 \\
\bottomrule
\end{tabular}
\caption{Parameter source analysis for the `use' and `pick\_up' actions.}
\label{tab:parameter_sources}
\end{table}
We quantify this predictability by modeling the sequence as a $k$-th order Markov chain and measuring the reduction in conditional entropy~\cite{shannon1948mathematical}. The analysis reveals a substantial drop in uncertainty: a 0-order model (assuming independence) yields a baseline entropy of 3.50 bits, this value decreases to 2.52 bits for a 1st-order model and drops further to 1.93 bits for a 2nd-order model.

To validate the statistical significance of these sequential dependencies, we performed likelihood ratio tests ($G^2$-tests, N = 6014). 
The results confirm that each increase in model order yields a highly significant improvement in fit. 
Specifically, the transitions from a 0-order to a 1st-order model, and from a 1st-order to a 2nd-order model, are both statistically significant, with $G^2(361) = 9390.70$ and $G^2(3914) = 5437.03$ respectively ($p < .001$ for both).

We provide a rigorous theoretical foundation for AutoTool in \textbf{Appendix~F}.
\begin{figure}[!] 
    \centering
    \includegraphics[width=\linewidth]{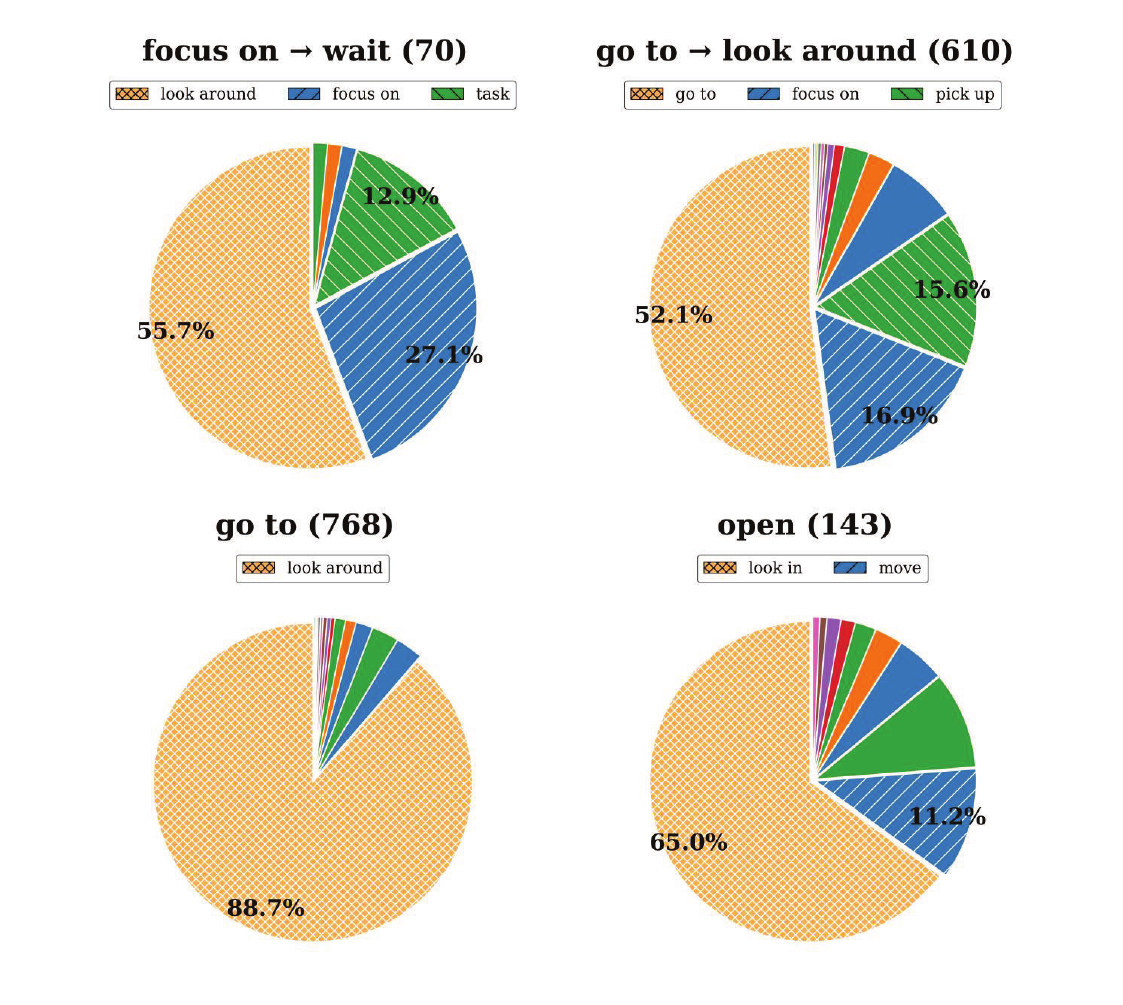}
    \caption{The distribution of successor tools for different tools or tool sequences.}
    \label{fig:successor}
\end{figure}

Empirically, this manifests as highly skewed successor distributions, as illustrated in Figure~\ref{fig:successor}. The \textit{go\_to} action is followed by \textit{look\_around} in 88.7\% of cases. Similarly, after the sequence \textit{focus\_on} → \textit{wait}, the next action is \textit{look\_around} with a high probability of 55.7\%.
These skewed distributions, where the next action often concentrates on one or two highly likely candidates, confirm the low-entropy nature of the task and demonstrate that non-LLM prediction is not only possible but highly feasible. Furthermore, the sources for directly transferred parameters are highly concentrated. As detailed in Table~\ref{tab:parameter_sources}, a narrow set of preceding actions often provides a substantial portion of the necessary inputs. For instance, the top source alone accounts for 44.8\% of parameters for \textit{use(target)} and 40.1\% for \textit{pick\_up(OBJ)}.

This quantifiable inertia---as revealed through theoretical and empirical analysis in ScienceWorld---forms the foundation for our proposed method.

\section{Methodology}

\subsection{Problem Statement}
\label{sec:problem statement}

In LLM-driven agent systems, an agent selects an action $a_t = (\text{tool}_{t+1}, \text{params}_{t+1})$ at each timestep $t$ based on an observation $o_t$, a task goal $G$, and a set of available tools $\mathcal{T}$. Current methods, such as ReAct, typically infer $a_t \sim p_{\text{LLM}}(a|o_t,G,\mathcal{T})$ via a complete LLM inference. This incurs significant computational costs, limiting their applicability in resource-constrained or real-time scenarios.

To address this limitation, we define our problem as follows:
Given a dataset of trajectories $D_{\text{hist}} = \{\tau_1, \tau_2, \ldots, \tau_N\}$ collected from historical task executions, where each trajectory $\tau_i = (o_0^{(i)}, a_0^{(i)}, o_1^{(i)}, a_1^{(i)}, \ldots)$ records a sequence of observations and actions.
Our objective is to construct $M_{\text{AutoTool}}$, a training-free decision-making algorithm based on the \emph{Tool Inertia Graph}, which can selectively bypass the LLM for predictable actions to improve efficiency.

\begin{figure*}[t]
    \centering
    \includegraphics[width=\textwidth, keepaspectratio]{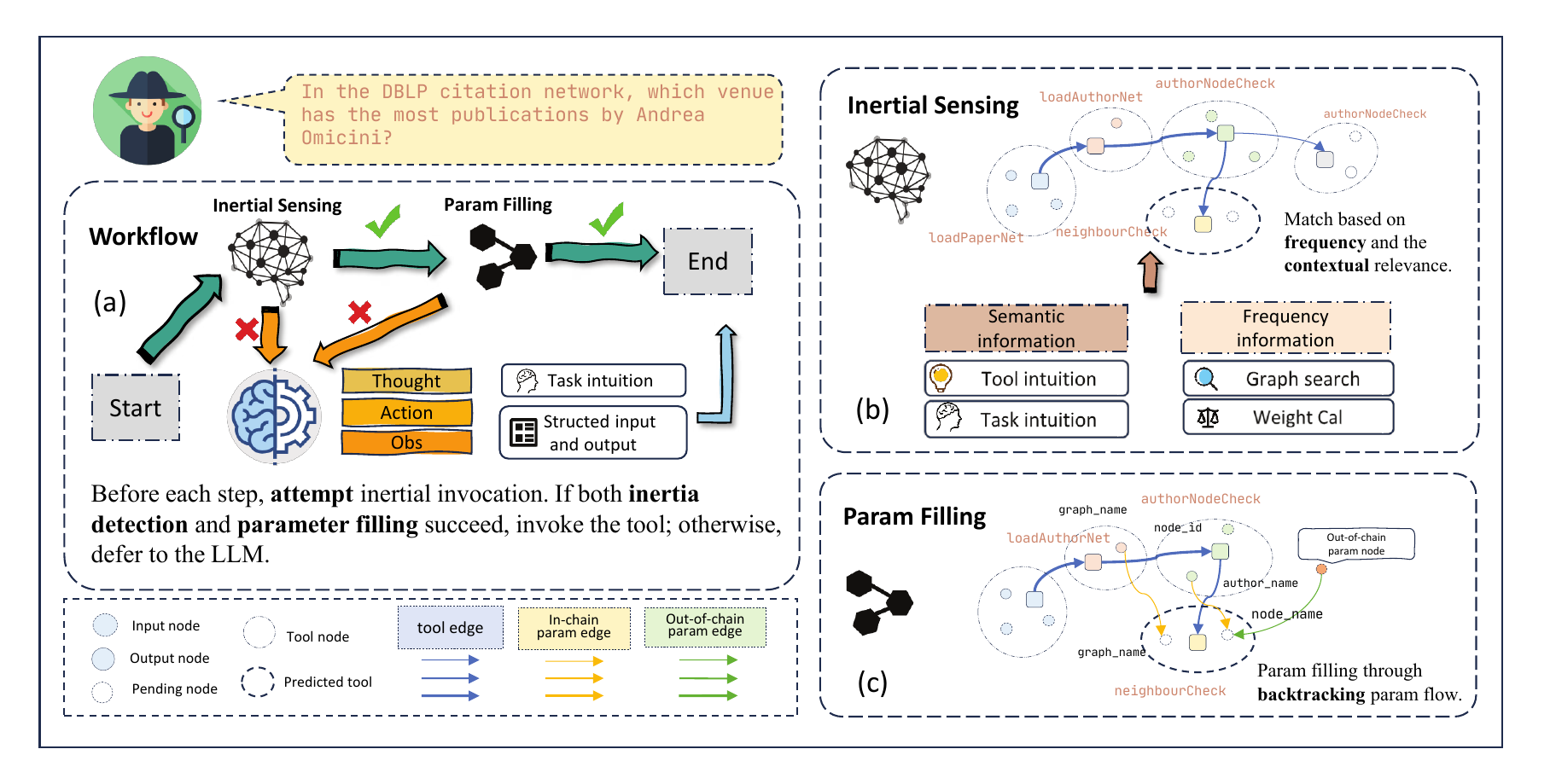}
    
    \caption{An overview of the AutoTool framework. This figure illustrates our proposed workflow, which initiates with the Inertia Sensing module predicting the next likely tool by exploiting historical usage patterns. If a high-confidence tool is identified, the Parameter Filling module then populates its required parameters. Only when both stages succeed is the tool executed directly via the inertial path, bypassing a costly LLM call. Otherwise, the system reverts to a standard LLM invocation.}
    
    \label{fig:workflow_wide}
\end{figure*}

\subsection{Overview}

Figure~\ref{fig:workflow_wide} illustrates the AutoTool framework, which attempts an `inertial invocation' before each standard LLM call to bypass costly inference. This process operates in two stages: first, the Inertia Sensing module (Fig.~\ref{fig:workflow_wide}b) predicts the next likely tool by combining historical frequency and contextual relevance. If a tool is identified, the Parameter Filling module (Fig.~\ref{fig:workflow_wide}c) then populates its arguments by backtracking the parameter flow on the graph. A tool is executed directly only if both stages succeed, thus bypassing a costly LLM inference. 
We introduce each module below and \textbf{defer more details and pseudo code of the algorithms to Appendix A due to page limit}.

\subsection{Graph Construction}

\label{sec:graph construction}

The \textbf{Tool Inertia Graph} (TIG) is a dynamic graph, denoted as $G_t=(V_t, E_t, W_t)$. It is incrementally constructed from execution trajectories and can also be bootstrapped with prior knowledge.
The TIG's node structure ($V_t$) is hierarchical:
\begin{itemize}
    \item \textbf{Tool Nodes} ($v_i$): Primary nodes, each representing an available tool, $tool_k \in \mathcal{T}$. They store the tool's functional description and track execution-level attributes, such as success or failure status. Notably, these nodes can be initially constructed from basic tool documentation alone, ensuring the graph's scalability. Each Tool Node encapsulates a dynamic subgraph containing:
        \begin{itemize}
            \item \textbf{Parameter Nodes} ($p_i$): Secondary nodes within the subgraph, each representing a specific input or output parameter of the parent Tool Node. 
        \end{itemize}
\end{itemize}
This hierarchical structure allows the TIG to meticulously track not only the sequence of tools but also the parameter-level data flows between them.

These nodes are connected by two types of directed edges ($E_t$):
\begin{itemize}
    \item \textbf{Tool Sequence Edges} ($e^{\text{tool}}_{ij}=(v_i, v_j)$): Connect Tool Nodes to represent sequential dependencies.
    \item \textbf{Parameter Dependency Edges} ($e^{\text{param}}_{xy}=(p_x, p_y)$): Connect Parameter Nodes to model the data flow between tools.
\end{itemize}

The initial creation and continuous update of these edges and their corresponding weights ($W_t$) are detailed in Section~\ref{sec:edge filling}. This online construction mechanism ensures the TIG promptly reflects the agent's latest experiences and learned patterns, making inertial predictions both adaptive and effective.

\subsection{Edge Filling}

\label{sec:edge filling}

The TIG learns and adapts from execution trajectories by dynamically creating and updating the edges that model both sequential tool dependencies and parameter-level data flows.

\paragraph{Tool Sequence Edges.}

When a tool $tool_j$ is executed immediately after $tool_i$, a Tool Sequence Edge ($e^{\text{tool}}_{ij}$) is created between them if it does not exist, typically with an initial weight of 1.
The edge's weight is reinforced exclusively by high-confidence sequences generated from the LLM. This design choice prevents error propagation from potentially sub-optimal inertial calls.

More critically than just recording frequencies, AutoTool enables the TIG to differentiate between effective and ineffective pathways. 
Each sequence edge is associated with a posterior efficacy score, which is continuously updated based on execution feedback. 
If an inertia-driven tool call is successful (judged by environmental feedback or task progression), the weight of the corresponding edge is incremented. Conversely, a failure penalizes the edge by decrementing its weight. This online learning mechanism allows the TIG not only to record historical co-occurrence frequencies but also to learn the \textit{actual effectiveness} of tool sequences, enabling robust adaptation to changing task dynamics.
\paragraph{Parameter Dependency Edges.}
Tracking parameter flow is crucial for automating parameter filling. When a tool is invoked, our framework parses its structured input and output. It then backtracks through the execution history to identify the source of each input parameter. When a parameter of the current tool inherits its value from any parameter of a preceding tool, a Parameter Dependency Edge ($e^{\text{param}}_{xy}$) is established or reinforced between the corresponding Parameter Nodes. These edges encode recurring data transfer patterns, laying the groundwork for efficient, non-LLM parameter filling.

\subsection{Graph Searching}

At each decision step, instead of immediately invoking the LLM, AutoTool first performs a graph search to attempt an inertial call. This process involves two sequential stages: tool selection and parameter filling.

\subsubsection{Tool Selection via CIPS}
\label{sec:tool_selection}

First, AutoTool searches the TIG to identify candidate tools that have historically followed the sequence of the most recent $k$ tools. For each candidate, we calculate a \textbf{Comprehensive Inertia Potential Score (CIPS)}, which balances historical patterns with the current task context:
\begin{equation} \label{eq:cips}
\text{CIPS} = (1-\alpha) \cdot \text{Score}_{\text{freq}} + \alpha \cdot \text{Score}_{\text{ctx}}
\end{equation}
The \textbf{Frequency Score} ($\text{Score}_{\text{freq}}$) quantifies historical usage patterns extracted from TIG edge weights, while the \textbf{Contextual Score} ($\text{Score}_{\text{ctx}}$) evaluates semantic alignment between the agent's current intuition and the candidate tool's description.
We select the tool $v^*$ achieving the highest CIPS. If this score surpasses the predefined threshold ($\text{CIPS}(v^*) > \theta_{\text{inertial}}$), the process proceeds to parameter filling. Otherwise, the inertial attempt aborts and control reverts to the standard LLM inference module.

\subsubsection{Hierarchical Parameter Filling}
\label{sec:param_filling}

If a tool candidate $v^*$ passes the threshold, AutoTool attempts to populate its required parameters using a hierarchical, non-LLM strategy that follows a strict priority order, ensuring that the most reliable information sources are always prioritized. 

The primary method is dependency backtracking, which traverses the parameter dependency edges in the TIG to find an input's source in a preceding tool's output. If this fails, the framework attempts environmental state matching, using key states maintained by the agent (e.g., current location). As a final non-LLM attempt, it resorts to heuristic filling based on the agent's current state or task goal. The inertial call is executed only if all required parameters are successfully populated through this hierarchy. If any parameter remains undetermined, the inertial call is aborted, and the decision-making process falls back to the LLM, ensuring that only high-confidence, fully specified actions are executed via inertia.

\section{Evaluation}

We present the primary experimental results below. Due to the page limit, \textbf{additional results including inertia analysis on ToolBench, case studies, dynamic analysis, and parameter filling accuracy are provided in the appendix.}

\subsection{Experimental Setup}

\paragraph{Datasets}

We evaluate AutoTool on three diverse and challenging multi-hop benchmarks to assess its generalizability across different domains. We begin with \textbf{Alfworld}~\cite{shridhar2020alfworld}, a classic text-based simulator for embodied household tasks. For more complex procedural tasks, we employ \textbf{ScienceWorld}~\cite{wang2022scienceworld}, where agents must follow logical sequences in scientific experiments. Finally, to evaluate multi-step API usage, we use \textbf{ToolQuery-Academic}~\cite{ma2024agentboard}, a benchmark from AgentBoard involving queries to an academic database. Together, these datasets provide a comprehensive test environment, covering challenges from physical reasoning and procedural following to structured API interactions.

\begin{table*}[t]
\centering
\small 
\renewcommand{\arraystretch}{1.2}
\setlength{\tabcolsep}{5pt} 

\begin{tabular}{@{}l
                S[table-format=1.3]
                S[table-format=4.0]
                S[table-format=4.0]
                S[table-format=2.1]
                S[table-format=1.3]
                S[table-format=4.0]
                S[table-format=4.0]
                S[table-format=2.1]
                S[table-format=1.3]
                S[table-format=4.0]
                S[table-format=3.0]
                S[table-format=2.1]
                @{}}
\toprule
\multirow{2}{*}{\textbf{Method}} & \multicolumn{4}{c}{\textbf{AlfWorld}} & \multicolumn{4}{c}{\textbf{ScienceWorld}} & \multicolumn{4}{c}{\textbf{Tool-Query-Academia}} \\
\cmidrule(lr){2-5} \cmidrule(lr){6-9} \cmidrule(lr){10-13}
& {\textbf{PR}} & {\textbf{tok-in}} & {\textbf{tok-out}} & {\textbf{LLMC}} & {\textbf{PR}} & {\textbf{tok-in}} & {\textbf{tok-out}} & {\textbf{LLMC}} & {\textbf{PR}} & {\textbf{tok-in}} & {\textbf{tok-out}} & {\textbf{LLMC}} \\
\midrule
ReAct & 0.394 & 6560 & 2310 & 24.1 & 0.716 & 9574 & 1072 & 23.3 & 0.901 & 1230 & 658 & 7.58 \\
\rowcolor{gray!15}
+ AutoTool & \bfseries 0.531 & \bfseries 4110 & \bfseries 804 & \bfseries 20.4 & 0.708 & \bfseries 7377 & \bfseries 758 & \bfseries 17.8 & 0.895 & \bfseries 1070 & 717 & \bfseries 6.32 \\
\textit{SpeedUp} & {---} & \textbf{\textit{1.60x}} & \textbf{\textit{2.87x}} & \textbf{\textit{1.18x}} & {---} & \textbf{\textit{1.30x}} & \textbf{\textit{1.41x}} & \textbf{\textit{1.31x}} & {---} & \textbf{\textit{1.15x}} & \textbf{\textit{0.92x}} & \textbf{\textit{1.20x}} \\
\midrule
Reflexion & 0.481 & 6813 & 2379 & 30.7 & 0.730 & 7282 & 1211 & 24.9 & 0.917 & 1680 & 680 & 8.85 \\
\rowcolor{gray!15}
+ AutoTool & 0.453 & \bfseries 5130 & \bfseries 1976 & \bfseries 23.7 & 0.712 & \bfseries 7842 & \bfseries 1012 & \bfseries 19.5 & \bfseries 0.923 & \bfseries 1260 & \bfseries 569 & \bfseries 7.05 \\
\textit{SpeedUp} & {---} & \textbf{\textit{1.33x}} & \textbf{\textit{1.20x}} & \textbf{\textit{1.29x}} & {---} & \textbf{\textit{0.93x}} & \textbf{\textit{1.20x}} & \textbf{\textit{1.28x}} & {---} & \textbf{\textit{1.33x}} & \textbf{\textit{1.19x}} & \textbf{\textit{1.26x}} \\
\bottomrule
\end{tabular}
\caption{Performance and resource consumption comparison of baseline methods with and without AutoTool. progress rate (PR) measures task accuracy, while SpeedUp row shows the cost reduction ratio.}
\label{tab:speedup}
\end{table*}

\paragraph{Evaluation Metrics}
Our evaluation focuses on two core aspects: task execution accuracy and efficiency.

To measure accuracy, we adopt the \textbf{progress rate} metric from the AgentBoard testing framework~\cite{ma2024agentboard}. AgentBoard manually defines key sub-goals for each task. It then calculates the progress rate by comparing the sequence of sub-goals an agent actually achieves against the predefined sequence. This metric serves as our primary indicator of an agent's ability to successfully complete tasks.

To evaluate task execution efficiency, we focus on two key metrics. We use the average number of \textbf{LLM calls} as a robust, hardware-agnostic proxy for temporal efficiency, since API latency is the primary runtime bottleneck. Concurrently, we measure the average \textbf{token consumption} to quantify the computational cost.

\paragraph{Baselines}

To our knowledge, there is no other published open-source work designed to improve the efficiency of tool selection in LLM agents without calling LLMs. Given this, and because our framework is designed for easy integration, we evaluate AutoTool by applying it to two foundational agent paradigms: ReAct~\citep{yao2023react} and Reflexion~\citep{shinn2023reflexion}. This approach allows us to demonstrate its effectiveness as an enhancement module.

ReAct, a foundational LLM agent paradigm, uses a ``Thought-Action-Observation" loop for multi-step reasoning and interaction. Reflexion enhances ReAct by introducing a self-reflection mechanism powered by an LLM-based evaluator. This evaluator assesses the agent's trajectories and generates feedback to prompt the agent to reflect on its errors. Our experiments mirror the original paper's heuristic self-reflection: reflection is triggered by failed tool calls or three consecutive identical observations, with the LLM-generated reflection added to the agent's memory.

\paragraph{Settings} All experiments are conducted on a server equipped with four Intel(R) Xeon(R) Gold 5117 CPUs and four NVIDIA Tesla V100-SXM2-32GB GPUs. We use Llama4-Scout-17b as the default model. The sampling temperature is consistently set to 0.

\subsection{Speedup}

% \todo{Quantitative results highlighting speedups relative to baselines. }

Our core objective is to verify that AutoTool can significantly reduce computational overhead (measured by the number of LLM calls and token consumption) while while maintaining comparable task performance (measured by progress rate).

Notably, the graph construction time is negligible, as analyzed in Section~\ref{sec:overhead}. We primarily assess efficiency improvement by comparing the average LLM calls and token consumption of AutoTool-enhanced agents (ReAct+AutoTool and Reflexion+AutoTool) with their original baselines. We use SimCSE~\cite{gao2021simcse} to calculate the contextual relevance score.

In our experimental setup, core AutoTool hyperparameters were tuned to achieve a 10-30\% reduction in LLM calls, with $\theta_{\text{inertial}} = 0.1$ controlling the inertia trigger threshold and $\alpha = 0.5$ weighting contextual relevance.
To maximize task progress, we implement a uniform constraint that limits inertia calls to no more than 30\% of the total operations and explicitly prohibits consecutive inertia calls. To ensure fair comparison, we keep all agent components and prompts consistent with their respective baselines.

\begin{table*}[t]
\centering
\small
\renewcommand{\arraystretch}{1.2}
\setlength{\tabcolsep}{5pt}

\begin{tabular}{@{}l l S[table-format=1.2] S[table-format=2.4] S[table-format=1.4] S[table-format=1.4] S[table-format=2.4] S[table-format=1.5, table-number-alignment=center]@{}}
\toprule
\multirow{2}{*}{\textbf{Method}} & \multirow{2}{*}{\textbf{Dataset}} & {\textbf{Init}} & \multicolumn{3}{c}{\textbf{Semantic Modules (s)}} & \textbf{Total} & {\textbf{Overhead \% of}} \\
\cmidrule(lr){3-3} \cmidrule(lr){4-6} \cmidrule(lr){7-7} \cmidrule(lr){8-8}
& & {(SimSCE)} & {Intuition Emb.} & {Tool Emb.} & {Similarity Comp.} & {\textbf{Overhead (s)}} & {\textbf{Total Task Time}} \\
\midrule
\multirow{3}{.8in}{\centering ReAct+ AutoTool}
 & AlfWorld     & 4.04 & 23.5793 & 3.0367 & 1.1830 & 31.84 & \textbf{1.21} \\ % 0.01207 * 100
 & ScienceWorld & 4.03 & 21.1180 & 1.0173 & 0.4775 & 26.64 & \textbf{1.38} \\ % 0.01375 * 100
 & Academic     & 3.76 &  6.1420 & 0.1704 & 0.0210 & 10.09 & \textbf{4.16} \\ % 0.04164 * 100
\midrule
\multirow{3}{.8in}{\centering Reflexion+ AutoTool}
 & AlfWorld     & 4.02 & 35.3731 & 7.2560 & 2.1583 & 48.81 & \textbf{1.72} \\
 & ScienceWorld & 4.07 & 26.6970 & 2.2535 & 0.7911 & 33.81 & \textbf{1.53} \\
 & Academic     & 3.77 &  5.1539 & 0.1105 & 0.0153 &  9.05 & \textbf{3.75} \\
\bottomrule
\multicolumn{8}{@{}p{\dimexpr\textwidth-2\tabcolsep}@{}}{\footnotesize \textbf{Note:} `Init (SimSCE)' refers to the one-time cost of loading the SimSCE model into memory.} \\
\end{tabular}
\caption{Time cost analysis of semantic modules in AutoTool. This table details the overhead from computing contextual relevance and its percentage of the total task execution time. All time values are in seconds (s).}
\label{tab:semantic_overhead}
\end{table*}

\begin{table*}[t]
\centering
\small
\renewcommand{\arraystretch}{1.2}
\setlength{\tabcolsep}{4pt}

\begin{tabular}{@{}l l S[table-format=1.4] S[table-format=1.4] S[table-format=1.4] S[table-format=1.5] S[table-format=1.4] S[table-format=4.1] S[table-format=4.0] S[table-format=4.0]@{}}
\toprule
\multirow{2}{*}{\textbf{Method}} & \multirow{2}{*}{\textbf{Dataset}} & \multicolumn{5}{c}{\textbf{AutoTool Core Modules (s)}} & {\textbf{LLM Time}} & \multicolumn{2}{c}{\textbf{Action Counts}} \\
\cmidrule(lr){3-7} \cmidrule(lr){8-8} \cmidrule(lr){9-10}
& & {Graph Const.} & {Graph Search} & {Parsing} & {Param Filling} & {Gen. Action} & {(s)} & {Inertial} & {Total} \\
\midrule
% --- ReAct+AutoTool Group ---
\multirow{3}{.8in}{\centering ReAct+ AutoTool}
 & AlfWorld     & 0.3159 & 1.5026 & 0.2370 & 0.4653 & 0.0267 & 2604.1 & 1076 & 3605 \\
 & ScienceWorld & 0.2414 & 0.9907 & 0.4578 & 0.1839 & 0.0192 & 1909.4 &  690 & 2316 \\
 & Academic     & 0.0286 & 0.0201 & 0.0218 & 0.0195 & 0.0031 &  232.1 &   49 &  203 \\
\midrule
% --- Reflexion+AutoTool Group ---
\multirow{3}{.8in}{\centering Reflexion+ AutoTool}
 & AlfWorld     & 0.4460 & 0.9933 & 0.2531 & 0.6226 & 0.0064 & 2799.8 & 1080 & 3763 \\
 & ScienceWorld & 0.3500 & 0.6857 & 0.5055 & 0.2914 & 0.0075 & 2173.8 &  689 & 2406 \\
 & Academic     & 0.0320 & 0.0139 & 0.0213 & 0.0119 & 0.0014 &  232.5 &   34 &  184 \\
\bottomrule
\multicolumn{10}{@{}p{\dimexpr\textwidth-2\tabcolsep}@{}}{\footnotesize \textbf{Note:} `LLM Time' is the LLM inference time within the AutoTool framework, while `Parsing' is the time spent parsing LLM outputs and tool outputs.} \\
\end{tabular}
\caption{Time cost and action count analysis of AutoTool's core modules. This table details the performance of non-semantic components within the AutoTool framework.}
\label{tab:core_modules_overhead}
\end{table*}

All experiments are conducted in a pure cold-start setting. The core graph is built online from scratch \textbf{without any prior trajectories}, ensuring fairness across comparisons.

As shown in Table~\ref{tab:speedup}, introducing AutoTool's optimization mechanism yields substantial efficiency gains across all datasets. On average, it reduces the LLM call count by 15\% to 25\% and the total token consumption by 10\% to 40\%.

\paragraph{For ReAct+AutoTool} We integrate into the TIG a specialized fault tolerance mechanism—designed as a preconfigured recovery path that activates upon detecting consecutive tool failures. It triggers a check operation to retrieve the current list of available tools, enabling the agent to re-orient itself and break out of ineffective exploration loops.

On the AlfWorld dataset, the effectiveness of this approach is particularly striking, as the TIG not only delivers substantial efficiency gains, with a 1.18x SpeedUp in LLM calls and over 1.60x in total token consumption, but also boosts progress rate, which in turn leads to fewer total execution steps. 
On the ScienceWorld and ToolQuery-Academic datasets, ReAct+AutoTool also demonstrates stable efficiency gains in both LLM call counts and token consumption, with total SpeedUp values of 1.3x/1.3x/1.4x (for tok-in, tok-out, and LLMC respectively) and 1.2x/1.2x/1.0x. In essence, the efficiency gains are attributable to the inertia graph's predictive capability, whereas the progress rate improvement (especially on AlfWorld) and overall stability are ensured by the fault-tolerance mechanism and the 30\% cap on inertial calls respectively.

\paragraph{For Reflexion+AutoTool} To avoid conflicting with its inherent reflection mechanism, we do not introduce a preconfigured recovery path. Despite this, AutoTool is still able to effectively utilize tool usage inertia and maintain competitive performance.
A comparison between Reflexion+AutoTool and ReAct+AutoTool reveals that the former yields a superior progress rate, primarily because the Reflexion mechanism improves the quality of collected inertia trajectories by reducing meaningless trial operations.

Furthermore, to assess the model-agnostic nature of our framework, we replicated experiments on ScienceWorld using diverse LLMs, including \textbf{Llama-3.3-70B}~\cite{dubey2024llama3}, \textbf{Qwen2.5-72B}~\cite{hui2024qwen2}, and \textbf{DeepSeekV3}~\cite{liu2024deepseek}. As detailed in \textbf{Appendix C}, AutoTool consistently delivered significant efficiency gains across all models, demonstrating that our method is a robust architectural improvement, not dependent on a specific model's characteristics.

\subsection{Overhead Analysis}

\label{sec:overhead}

To quantify the computational overhead introduced by our framework, we systematically analyzed the time consumption of its core components for both ReAct+AutoTool and Reflexion+AutoTool across all three datasets. We dissect this overhead into two primary categories: the non-semantic modules (e.g., graph search and construction), detailed in Table~\ref{tab:core_modules_overhead}, and the semantic relevance calculation modules, which constitute the primary overhead, presented in Table~\ref{tab:semantic_overhead}.

As detailed in Table~\ref{tab:core_modules_overhead}, our analysis reveals that the overhead from AutoTool’s non-semantic components is minimal, typically remaining on the order of seconds even when LLM inference time for a task exceeds a thousand seconds. The primary overhead stems from contextual relevance calculation, which involves embedding both the agent's intuition and tool descriptions. However, as Table~\ref{tab:semantic_overhead} shows, this cost is negligible: computing contextual relevance accounts for only 2.7\% ± 1.5\% of total task execution time, a mere fraction of standard LLM inference.

\begin{table}[t]
\centering
\small
\renewcommand{\arraystretch}{1.15}
\setlength{\tabcolsep}{4pt}

\begin{tabular}{@{}
    c@{\hspace{0.8em}} 
    c@{\hspace{0.8em}}
    S[table-format=1.3, table-space-text-post=\bfseries]@{\hspace{0.8em}}
    S[table-format=5.0]@{\hspace{0.8em}}
    S[table-format=3.0]@{\hspace{0.8em}}
    S[table-format=2.2, table-space-text-post=\bfseries]@{\hspace{0.8em}} 
    S[table-format=2.2, table-space-text-post=\bfseries]
    @{}}
\toprule
\textbf{\(\theta_{in}\)} & \textbf{\(\alpha\)} & {\textbf{PR}} & {\textbf{Tok-In}} & {\textbf{Tok-Out}} & {\textbf{LLMC}} & {\textbf{Avg Act}} \\
\midrule
% --- Group 1: theta = 0.1 ---
\multirow{3}{*}{0.1} 
 & 0.3 & \textbf{0.719} & \textbf{67724} & 710 & 17.13 & 24.45 \\
 & 0.5 & 0.694 & 69010 & \textbf{702} & \bfseries \textbf{16.85} & \bfseries \textbf{24.18} \\ 
 & 0.7 & 0.686 & 73897 & 785 & 18.38 & 26.47 \\
\midrule
% --- Group 2: theta = 0.15 ---
\multirow{3}{*}{0.15}
 & 0.3 & 0.706 & 71077 & 765 & 17.72 & 25.36 \\
 & 0.5 & \bfseries 0.715 & 70640 & 744 & 17.66 & 25.22 \\
 & 0.7 & 0.687 & 69287 & 726 & 17.33 & 24.75 \\
\midrule
% --- Group 3: theta = 0.2 ---
\multirow{3}{*}{0.2}  
 & 0.3 & 0.696 & 78213 & 812 & 19.14 & 26.93 \\
 & 0.5 & 0.695 & 73515 & 854 & 17.94 & 24.88 \\
 & 0.7 & \bfseries 0.696 & 73846 & 861 & 18.12 & 24.67 \\
\bottomrule
\end{tabular}
\caption{Sensitivity analysis of AutoTool on the ScienceWorld dataset. We vary the inertia trigger threshold \(\theta_{in}\) and the contextual relevance weight \(\alpha\). }
\label{tab:sensitivity_analysis}
\end{table}

\subsection{Sensitivity Analysis}
To investigate the impact of AutoTool's key hyperparameters on its performance, we conduct a sensitivity analysis on the inertia trigger threshold \(\theta_{inertial}\) and the semantic similarity weight $\alpha$. The experiments are performed on the ScienceWorld dataset with ReAct+AutoTool, and the results are shown in Table \ref{tab:sensitivity_analysis}. The experimental results largely meet our expectations: a lower \(\theta_{inertial}\)  (e.g., 0.1) tends to trigger more inertia calls, thereby most effectively reducing the average number of LLM calls (e.g., reaching the lowest value of 16.85 among all tested combinations when $\alpha = 0.5$) and the corresponding token consumption.

Interestingly, the progress rate remains stable despite the lower threshold. We attribute this to the two key constraints: a 30\% cap on total inertia calls, and a prohibition on consecutive ones. Indeed, even with a relatively high \(\theta_{inertial}\) of 0.2, the number of triggered inertia calls already approaches or reaches this 30\% ceiling. This reveals an insight into our design: under these constraints, a more lenient \(\theta_{inertial}\) is not only safe but beneficial. It allows AutoTool to capture a more diverse range of effective inertia patterns without being overwhelmed by low-quality calls, thus avoiding the rigidity and insufficiency of an overly strict threshold.

\section{Conclusion}
We propose AutoTool, a graph-based and lightweight tool selection framework. Rather than simply following observed tool usage inertia, AutoTool treats it as a behavior to be actively managed. By innovatively integrating statistical structure into the design of LLM agents, AutoTool leverages the observed tool usage inertia to effectively address the high latency and resource consumption associated with multi-step tool selection in existing frameworks. While we detail the framework's current limitations in \textbf{Appendix~G}, our method has demonstrated strong generalization and adaptability, showing great potential for practical applications.

\section*{Acknowledgments}

This work is supported by the National Natural Science Foundation of China (Grant No. 62502174).

\bibliography{aaai2026}
\appendix
\setcounter{secnumdepth}{1}
\newpage
\input{appendix}
\end{document}

%% file: appendix.tex
\section{Details of AutoTool}
\label{sec:alg_detail}
\subsection{Data Structure}
The core data structure of AutoTool is the Tool Inertia Graph (TIG), designed to capture and leverage sequential patterns in tool usage and parameter dependencies. The TIG is primarily composed of several interconnected components that work in concert.

At the foundational level are \textbf{ToolNodes}, each precisely representing an available tool within the system. A ToolNode encapsulates the tool's name, functional description, and a formalized list of its input arguments (args) and output arguments (returns). To manage data flow within individual tools more granularly, each ToolNode embeds a \textbf{ParamGraph}. This dedicated subgraph is responsible for maintaining \textbf{ParamNodes} associated with the tool. These ParamNodes not only distinguish between input and output attributes but also cache example values observed during runtime, thereby providing a basis for subsequent parameter filling.

To track the transfer of data across different tool invocations, the TIG incorporates \textbf{ParamEdges}. This constitutes a core mapping structure, conceptually organized as \texttt{\{target\_tool: \{target\_param: \{(source\_tool, source\_param): ParamEdge\_instance\}\}\}}. Each individual \texttt{ParamEdge\_instance} meticulously records a specific historical data flow: it signifies that an output value or input value from a particular ``source\_param" of a ``source\_tool" has previously served as an input for a specific ``target\_param" of a subsequent ``target\_tool." A counter within the ParamEdge quantifies the strength and frequency of this observed dependency.

Complementing the node and data-flow structures, the TIG maintains \textbf{ToolPaths} to learn and utilize sequential patterns of tool invocations. Each ToolPath object stores an observed complete sequence of tool calls (represented as a list of tool names) along with the frequency of that specific sequence's occurrence. For erroneous tool invocations, AutoTool simply add a tool call chain with a negative frequency, but ensure that its length is greater than the inertia observation window.

For efficient retrieval and management of these paths, the TIG internally employs several key mappings. \texttt{self.path\_index} provides a rapid index from an individual tool name to a set of all ToolPath IDs that contain that tool. The \texttt{self.paths} list allows direct access to the ToolPath object itself (containing the tool sequence, frequency, etc.) via its unique ID. Furthermore, \texttt{self.paths\_lookup} enables quick lookup of a ToolPath ID from a tuple representation of a tool sequence, which is crucial for efficiently checking path existence and updating path frequencies.

Collectively, these components—ToolNodes with their embedded ParamGraphs, the graph of ParamEdges, and the systematically managed ToolPaths—form the foundation of the TIG. This architecture enables the TIG to dynamically learn from execution history and effectively support inertia-driven tool selection.

\subsection{Tool Prediction}

AutoTool predicts the next tool by analyzing a segment of recently executed tools called the \textbf{current\_sequence}. This current\_sequence is a truncated portion of the full tool execution history, and its length is defined by a variable called the \emph{inertia window}. The inertia window is typically set to 2, this means AutoTool rely on two consecutive action sequences for inertia prediction. When the length is 1, the lack of contextual information results in lower accuracy for inertia prediction. Increasing the length to 3 or more theoretically allows for capturing longer dependencies, but on the current datasets, the number of historical paths that meet the criteria significantly decreases, leading to a decline in the model's generalization ability and an increased risk of overfitting to specific long sequences.
Exploring a dynamically changing inertia window is a direction for further optimization.

\begin{algorithm}[htbp]
  \SetAlgoLined
  \caption{ToolPrediction}
  \label{algo:inertial_prediction}
  \SetKwInOut{Input}{Input}
  \SetKwInOut{Output}{Output}
  \Input{\emph{current\_history}, \emph{current\_memory}, \emph{tool\_graph}}
  \Output{\emph{predicted\_tool} (String or None), \emph{confidence\_score} (Float)}

  \BlankLine
  \tcp{Goal: Predict the next tool based on historical patterns and current context.}
  \emph{current\_sequence} := GetRecentToolSequence(\emph{current\_history})\;
  \emph{intuition} := ExtractRelevantThought(\emph{current\_memory})\;

  \BlankLine
  \tcp{Step 1: Candidate Path Identification}
  \emph{candidate\_paths} := FindMatchingPaths(\emph{current\_sequence}, \emph{tool\_graph})\;
  \If{\emph{candidate\_paths} is empty}{
    \KwRet{None, 0.0}\;
  }

  \BlankLine
  \tcp{Step 2: Scoring Candidate Next Tools}
  \emph{scored\_next\_tools} := EmptyMap\;
  \ForEach{path \KwSty{in} \emph{candidate\_paths}}{
    \emph{next\_tool} := GetNextToolInPath(\emph{path}, \emph{current\_sequence})\;
    \If{\emph{next\_tool} exists}{
      \emph{freq\_score} := CalculateFrequencyScore(\emph{next\_tool}, \emph{path}, \emph{tool\_graph})\;
      \emph{sem\_score} := CalculateSemanticSimilarity(\emph{next\_tool}'s description, \emph{intuition})\;
      \emph{combined\_score} := WeightedSum(\emph{freq\_score}, \emph{sem\_score})\;
      UpdateOrAdd(\emph{scored\_next\_tools}, \emph{next\_tool}, \emph{combined\_score})\;
    }
  }
  \If{\emph{scored\_next\_tools} is empty}{
    \KwRet{None, 0.0}\;
  }

  \BlankLine
  \tcp{Step 3: Prediction Decision}
  \emph{best\_tool}, \emph{highest\_score} := GetToolWithMaxScore(\emph{scored\_next\_tools})\;
  \KwRet{\emph{best\_tool}, \emph{highest\_score}}\;

\end{algorithm}

The graph searching process begins by identifying candidate historical \texttt{ToolPaths} that
contain the \texttt{current\_sequence} as a sub-sequence. Tool Prediction module first use \texttt{TIG.path\_index} to get the set of path IDs associated with the first tool in \texttt{current\_sequence}. Then, for each subsequent
tool in \texttt{current\_sequence}, this set of path IDs is intersected with 
the set of path IDs associated with that tool. This efficiently narrows down
to \texttt{potential\_path\_indices} that are supersets of the current execution window.

Once these \texttt{potential\_path\_indices} are obtained, Tool Prediction module iterate through each corresponding \texttt{ToolPath}. It first verifies whether the \texttt{current\_sequence} is indeed a sub-sequence of the historical path. Then it checks if there is at least one tool following the \texttt{current\_sequence} in that historical path. If both conditions are satisfied, the tool that immediately follows the \texttt{current\_sequence} in the historical path is considered a candidate for the next action. The weight for each unique candidate next tool in the \texttt{candidate\_tool\_dic} is then updated based on the \texttt{frequency} attribute of the historical \texttt{ToolPath}, denoted as $w(c_j)$.

The Inertia Confidence Score (ICS) calculation often yields multiple candidate tools for the next action. AutoTool first computes the total aggregated weight $W_{\text{total}} = \sum w(c_i)$ across all candidate tools $c_i$, reflecting the overall historical evidence strength for any known subsequent tool. To prevent premature high-confidence inertial calls when historical data is sparse (i.e., $W_{\text{total}}$ is small), an Inertia Confidence Factor (ICF) is introduced, derived by applying a slowly increasing, bounded-by-1 exponential function to $W_{\text{total}}$: $\text{ICF} = 1 - k^{-W_{\text{total}}}$, where $k$ is a constant greater than 1 (e.g., $k=1.1$). When $W_{\text{total}}$ is small, ICF approaches 0, suppressing overall confidence; as $W_{\text{total}}$ grows, ICF approaches 1, indicating higher confidence. Finally, the Inertia Confidence Score (ICS) for each candidate tool $c_j$ is calculated as: $\text{ICS}(c_j) = \left( \frac{w(c_j)}{W_{\text{total}}} \right) \times \text{ICF}$. This score is then combined with contextual relevance to form the final comprehensive inertia potential score. This mechanism ensures AutoTool effectively leverages inertia when sufficient data is available while maintaining conservative decision-making in data-scarce scenarios.

\subsection{Param Filling}

A significant challenge in designing a general-purpose parameter filling mechanism arises from the considerable heterogeneity across different datasets, particularly concerning the input/output structures of tools and varying environment states. To ensure broad applicability and maintain a modular architecture, AutoTool adopts an \textbf{Adaptor pattern}. This approach effectively decouples the core parameter filling framework from dataset-specific intricacies.

For each new dataset or environment, a corresponding concrete Adaptor subclass must be implemented to bridge these specific details with the generic filling logic. The core framework then orchestrates the Adaptor and other components to perform the comprehensive parameter filling process. The base class definition for the Adaptor interface is presented here.

\begin{lstlisting}[caption={Abstract Base Class for Environment Adapters}, label={lst:env_adapter}]
import abc 
from typing import Dict, List, Any, Tuple, Optional, Set 

class EnvironmentAdapter(abc.ABC):
    """
    Environment adapter base class, defines interfaces related to specific environments/datasets.
    Each new environment should implement a concrete adapter subclass.
    """

    def __init__(self, debug: bool = False):
        """
        Initialize the environment adapter
        Args:
            debug: Whether to enable debug mode
        """
        self.debug = debug

    @abc.abstractmethod
    def reset(self, init_observation: str = None):
        pass

    @abc.abstractmethod
    def parse_action(self, action_text: str, tool_descriptions: Dict) -> Optional[Dict[str, Any]]:
        """
        Parse action text and return structured action data
        Args:
            action_text: Action text
            tool_descriptions: Tool description dictionary
        Returns:
            Parsed action dictionary, format: {"tool_name": str, "inputs": Dict[str, Any]}
        """
        pass

    @abc.abstractmethod
    def infer_output(self, tool_name: str, inputs: Dict[str, Any], result: Any) -> Dict[str, Any]:
        """
        Infer structured output based on tool execution result
        Args:
            tool_name: Tool name
            inputs: Input parameters
            result: Execution result
        Returns:
            Inferred structured output dictionary
        """
        pass

    @abc.abstractmethod
    def update_state(self, action_parsed: Dict[str, Any], structured_outputs: Dict[str, Any]) -> None:
        """
        Update environment state based on action and output
        Args:
            action_parsed: Parsed action dictionary
            structured_outputs: Structured output dictionary
        """
        pass

    @abc.abstractmethod
    def get_contextual_params(self, action_type: str, missing_params: Set[str], required_params_info: Dict) -> Dict[str, Any]:
        """
        Infer parameter values based on current environment state
        Args:
            action_type: Target action type
            missing_params: Set of missing parameters
            required_params_info: Dictionary of required parameter descriptions
        Returns:
            Inferred parameter dictionary
        """
        pass

    @abc.abstractmethod
    def generate_action_from_params(self, action_type: str, params: Dict[str, Any]) -> str:
        pass
\end{lstlisting}

\begin{algorithm}[htbp]

  \SetAlgoLined
  \caption{ParameterFilling}
  \label{algo:inertial_filling}

  \SetKwInOut{Input}{Input}
  \SetKwInOut{Output}{Output}
  \Input{\emph{target\_tool}, \emph{exec\_history}, \emph{param\_dep\_graph}, \emph{env\_adapter}, \emph{tool\_graph}}
  \Output{\emph{filled\_parameters} (Map), \emph{was\_successful} (Boolean)}

  \BlankLine
  \tcp{Goal: Automatically populate input parameters for the \emph{target\_tool}.}
  \emph{required\_params} := GetInputSchema(\emph{target\_tool}, \emph{tool\_graph\_schema})\;
  \emph{filled\_params} := EmptyMap\;

  \BlankLine
  \tcp{Priority 1: Parameter Dependency Graph (Learned direct output\text{-}to\text{-}input links)}
  \ForEach{param\_name \KwSty{in} \emph{required\_params}}{
    \If{\emph{param\_name} not yet filled}{
      \emph{sources} := QuerySources(\emph{param\_dep\_graph}, \emph{target\_tool}, \emph{param\_name})\;
      \If{value found \KwSty{in} \emph{sources}}{
        \emph{value} := GetValueFromHistory(\emph{exec\_history}, source\_tool, source\_param)\;
        \If{\emph{value} is valid and type\text{-}compatible}{
          \emph{filled\_params}[\emph{param\_name}] := \emph{value}\;
        }
      }
    }
  }

  \BlankLine
  \tcp{Priority 2: Environment Context}
  \ForEach{param\_name \KwSty{in} \emph{required\_params}}{
    \If{\emph{param\_name} not yet filled}{
      \emph{value} := \emph{env\_adapter}.GetContextualValue(\emph{target\_tool}, \emph{param\_name})\;
      \If{\emph{value} is available and valid}{
        \emph{filled\_params}[\emph{param\_name}] := \emph{value}\;
      }
    }
  }
  \BlankLine
  \emph{all\_critical\_filled} := CheckIfAllCriticalParamsAreFilled(\emph{filled\_params}, \emph{required\_params})\;
  \KwRet{\emph{filled\_params}, \emph{all\_critical\_filled}}\;

\end{algorithm}

Future work will explore applying some prompt engineering to achieve a more structured presentation in status and tool output observation, which will make parameter filling adaptable to different datasets and environments much more easily, while minimizing the introduction of additional token overhead.

When a source parameter value is retrieved from historical data via backtracking and matching, AutoTool handles it based on its type. If the source value is singular (e.g., a string or number), it is directly assigned to the target parameter. However, a challenge arises when the source value is a list of multiple elements while the target parameter typically expects a single value, necessitating a selection strategy. While an ideal strategy would deeply integrate tool semantics and current task context, this often results in dataset-specific solutions that compromise generality. Although sophisticated, scenario-specific heuristic rules could be developed, our current AutoTool implementation employs a more general method to balance efficacy with simplicity: the framework maintains a record of all values already used for parameter filling within the ongoing task. When sourcing from a list, AutoTool randomly selects an element from that source list that has not yet appeared in this record. This approach provides a foundational mechanism for diversifying parameter inputs from list-type sources within the same task, acknowledging that more contextually-aware selection strategies represent an avenue for future enhancement.

\section{Macro-Level Inertia Analysis on ToolBench}

\begin{figure}[h!]
    \centering
    \includegraphics[width=\linewidth]{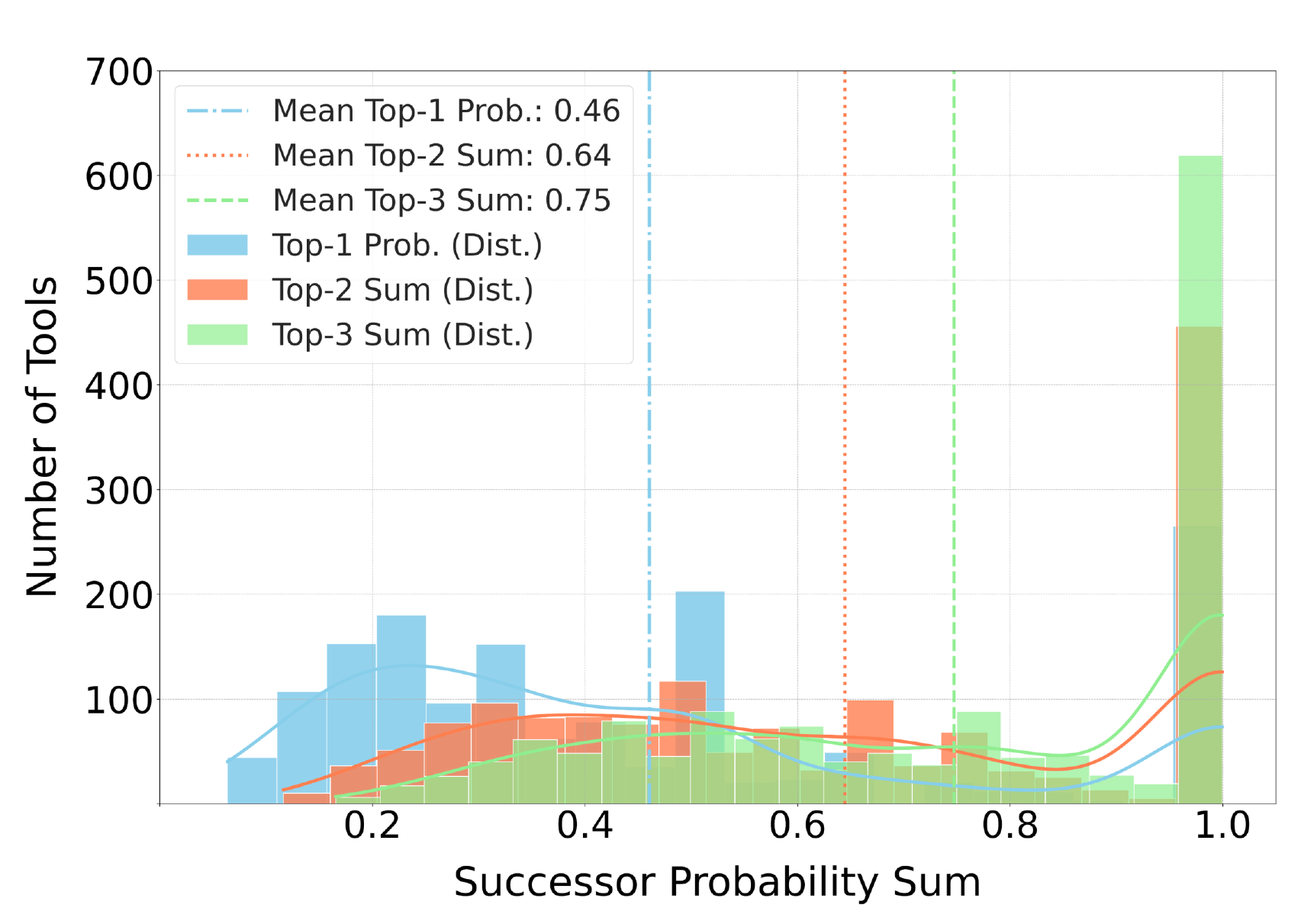} 
    \caption{ToolBench Inertia Analysis. We calculated the frequency distribution of successor tools for each tool and derived the proportion of its top-$k$ successor tools among all its successor tools.}
    \vspace{-10pt}
    \label{fig:toolbenchinertia}
\end{figure}

To validate the generalizability of our "Sequential Inertia" hypothesis beyond task-oriented domains, we conducted a supplementary macro-level analysis on the ToolBench G3 (Figure~\ref{fig:toolbenchinertia}). ToolBench is a large-scale benchmark featuring over 16,000 real-world APIs, with its G3 subset being a comprehensive collection of over 15,000 diverse tools. This dataset provides an ideal testbed for evaluating the structural properties of tool-calling behavior in a more open-ended, less constrained environment.

We processed 15,980 valid invocation trajectories from the G3 dataset. To ensure the analysis reflects meaningful sequential patterns, we applied a preprocessing step to remove repetitive tool calls (e.g., A → A) and overly short trajectories (length less than 2). Based on the cleaned trajectories, we constructed a transition probability matrix representing the likelihood of transitioning from any given tool to another.

Our primary metric for this macro-level analysis is the \textbf{system-wide conditional entropy}, which quantifies the average uncertainty of predicting the next tool, given the current one. The results are as follows:
\begin{itemize}
    \item \textbf{Calculated Conditional Entropy \textit{H(Y|X)}}: Our analysis revealed a system-level conditional entropy of only \textbf{3.62 bits}.

    \item \textbf{Maximum Theoretical Entropy (Uniform Baseline)}: For a system with 1,595 unique tools, the maximum possible entropy, assuming completely random transitions, is $\log_{2}(1595) \approx \textbf{10.64 bits}$.

    \item \textbf{Entropy Reduction}: The observed entropy of 3.62 bits represents a \textbf{65.96\% reduction} from the theoretical maximum.
\end{itemize}

The nearly 66\% reduction in entropy provides a powerful, formal guarantee of the system's non-random and highly predictable structure. It confirms that even in a vast and diverse ecosystem of real-world APIs like ToolBench, tool invocation is not a random walk. Instead, it is governed by strong, underlying patterns of usage. This macro-level evidence strongly supports our central claim that tool-calling inertia is a fundamental and generalizable property of LLM agents, making our AutoTool approach broadly applicable.

\section{Model Robustness}

To verify the generalization ability of our method across different models, we conducted experiments on the ScienceWorld dataset using a diverse set of models with varying parameter sizes and architectures. Specifically, we tested the following models: Llama-4-Scout-17B-16E-Instruct, DeepSeekV3, Llama3.3-70B-Instruction, and Qwen2.5-72B-Instruct-Turbo. The results are presented in Table \ref{tab:llm_robustness}.

Our method demonstrates strong generalization across different models. Despite significant performance differences among these models on the same dataset, our approach consistently reduces token consumption and the number of LLM calls while maintaining a comparable progress rate. This provides empirical evidence for further generalization in multiple scenarios.

\begin{table*}[htbp]
  \centering
  \caption{Performance Comparison of AutoTool with ReAct across Different LLMs on ScienceWorld. Metrics include Progress Rate (PR), Token Consumption (token-in, token-out), and Number of LLM Calls.}
  \label{tab:llm_robustness}
  \small 
  \setlength{\tabcolsep}{5pt} 
  \renewcommand{\arraystretch}{1.1}
  \begin{tabular}{@{}l l S[table-format=1.4, table-align-uncertainty=true, table-figures-uncertainty=1] S[table-format=5.2, table-align-uncertainty=true, table-figures-uncertainty=1] S[table-format=4.2, table-align-uncertainty=true, table-figures-uncertainty=1] S[table-format=2.2, table-align-uncertainty=true, table-figures-uncertainty=1]@{}}
    \toprule
    \textbf{Base LLM} & \textbf{Method} & {\textbf{Progress Rate}} & {\textbf{token-in}} & {\textbf{token-out}} & {\textbf{llm\_call}} \\
    \midrule
    \multirow{2}{*}{Llama-4-Scout-17B-16E-Instruct}
    & ReAct           & 0.7159 & 95743.62 & 1072.03 & 23.31 \\
    & ReAct+AutoTool  & 0.7082 & 73779.72 &  758.87 & 17.81 \\
    \midrule
    \multirow{2}{*}{Llama-3.3-70B-Instruct-Turbo}
    & ReAct           & 0.8126 & 90001.91 & 1687.08 & 19.15 \\
    & ReAct+AutoTool  & 0.7891 & 63886.16 & 1130.70 & 14.60 \\
    \midrule
    \multirow{2}{*}{Qwen2.5-72B-Instruct-Turbo}
    & ReAct           & 0.7600 & 81466.92 & 1168.15 & 19.71 \\
    & ReAct+AutoTool  & 0.7484 & 66563.08 &  876.54 & 15.93 \\ 
    \midrule 
    \multirow{2}{*}{DeepSeekV3}
    & ReAct           & 0.4479 & 112062.43 & 2204.80 & 25.00 \\ 
    & ReAct+AutoTool  & 0.5889 & 72608.53 & 871.43 & 17.26 \\ 
    \bottomrule
  \end{tabular}
\end{table*}

For DeepSeekV3, the poor instruction-following ability of the base model leads to a significant waste of solution steps as it fails to output action in the prescribed format. However, our preconfigured recovery path (two consecutive erroneous calls force a call to \emph{check valid action}, to verify the set of available tools) significantly improves the progress rate while reducing token consumption. Similar experimental phenomena were also observed in the test results of ReAct + AutoTool on alfworld, as detailed in Table \ref{tab:speedup}.

\section{Case Study}
To provide a comprehensive and balanced evaluation of AutoTool, this section presents both a detailed success case and an analysis of characteristic failure scenarios. This approach allows us to not only demonstrate the practical benefits of our method but also to transparently define its operational boundaries.

\subsection{Success Case Analysis}
\begin{figure}[htbp]
    \centering
    \includegraphics[width=\linewidth]{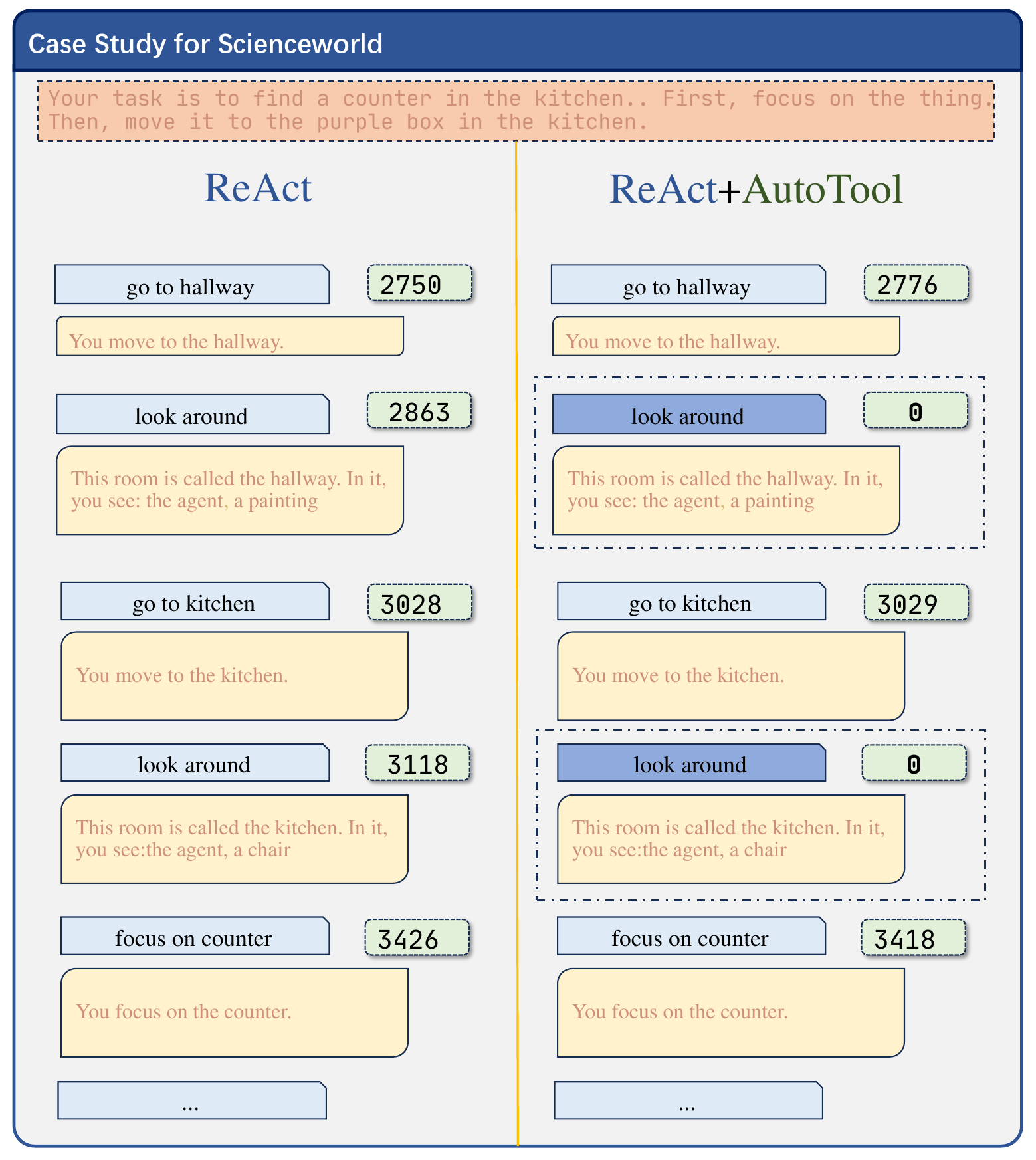}
    \caption{Case study for scienceworld}
    \label{fig:case_study}
\end{figure}

Here we present case studies comparing the ReAct agent with the AutoTool+ReAct agent. We illustrate examples from the 30th Scienceworld task in Figure \ref{fig:case_study}.

As shown on the right, \emph{Look around} is executed twice by inertial tool calling, which significantly reduces token consumption. Meanwhile, the baseline has to take one step at a time. This obvious inertial calling is harmless for the agent to solve the task. Even if the inertial calling leads to inefficient or erroneous actions, it barely causes any waste of tokens and time.

\subsection{Failure Case Analysis}
In the typical usage scenarios of AutoTool, there are three classic failure scenarios. An analysis of these scenarios not only reveals the methodological boundaries of AutoTool but also inversely demonstrates the necessity of our designed feedback and fallback mechanisms. All cases presented below are derived from authentic experimental logs.

\subsubsection{Scenario 1: Parameter Filling with Incorrect Context}
When an agent needs to fill parameters for an action, AutoTool might retrieve a ``frequently used" parameter from a completely unrelated historical memory, entirely ignoring the agent's current physical location and task constraints. This leads to action failure due to unsatisfied preconditions.

\begin{figure}[htbp]
    \centering
    \includegraphics[width=\linewidth]{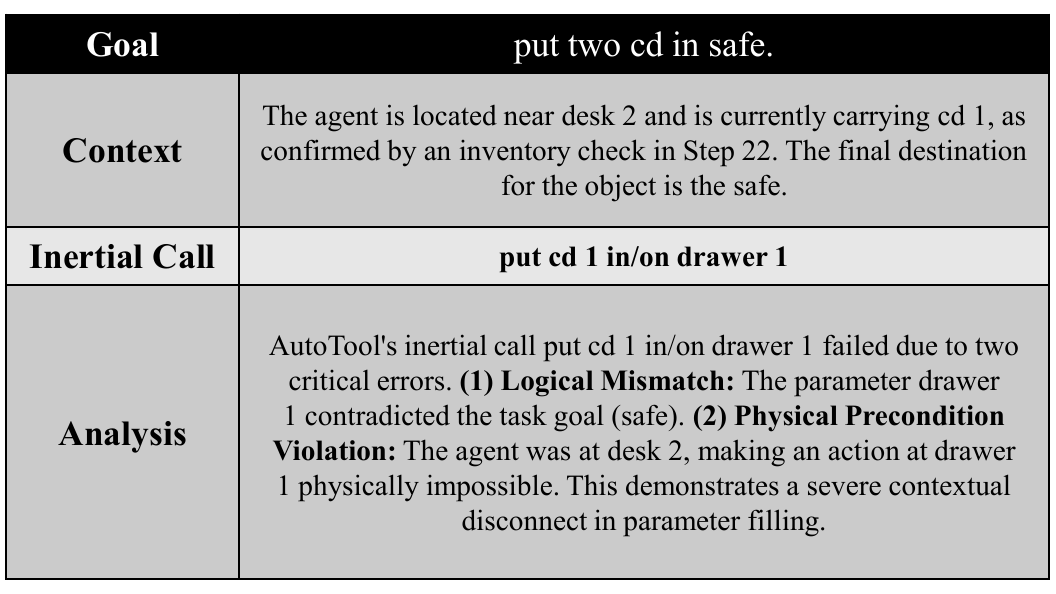} 
    \caption{Failure Scenario 1}
    \label{fig:case1}
\end{figure}

\subsubsection{Scenario 2: Inertia-Induced Redundant Actions}

A more subtle risk of inertia is that it can cause the agent to ``spin its wheels." If an action sequence is historically chained, AutoTool might mechanically execute the next step without first evaluating whether the current state already satisfies the desired condition. While this may not cause a task-level ``failure," it generates useless steps, wasting both time and execution turns.

\begin{figure}[htbp]
    \centering
    \includegraphics[width=\linewidth]{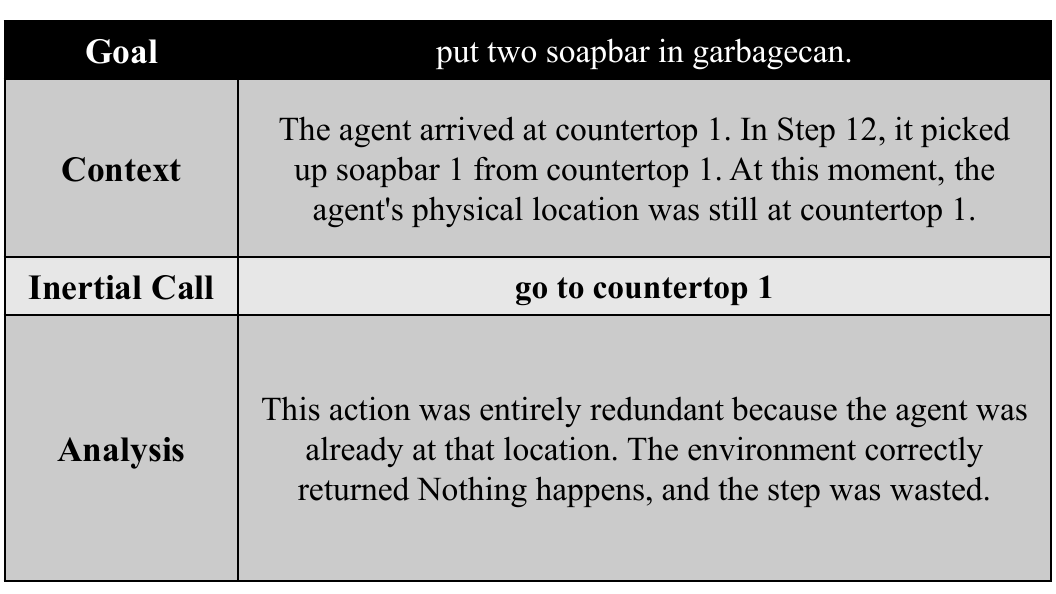}
    \caption{Failure Scenario 2}
    \label{fig:case2}
\end{figure}

\subsubsection{Scenario 3: Overgeneralization of Tool Usage Patterns}

AutoTool builds its inertia graph by learning high-frequency action sequences. However, it might learn a general pattern (e.g., go to $\rightarrow$ open)
 without understanding the semantic constraints of the objects to which it applies. A failure of "overgeneralization" occurs when it applies this general pattern to an object that does not support the action.

\begin{figure}[htbp] 
    \centering
    \includegraphics[width=\linewidth]{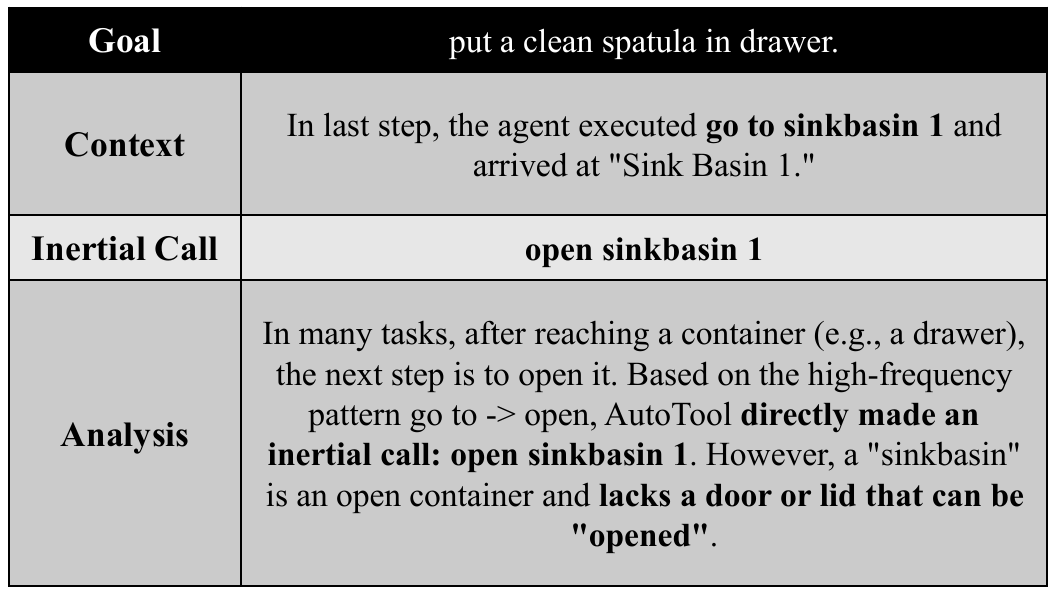}
    \caption{Failure Scenario 3}
    \label{fig:case3}
\end{figure}

\section{Dynamic Analysis}
\begin{figure*}[t] 
    \centering
    \begin{subfigure}[b]{0.32\textwidth}
        \centering
        \includegraphics[width=\textwidth]{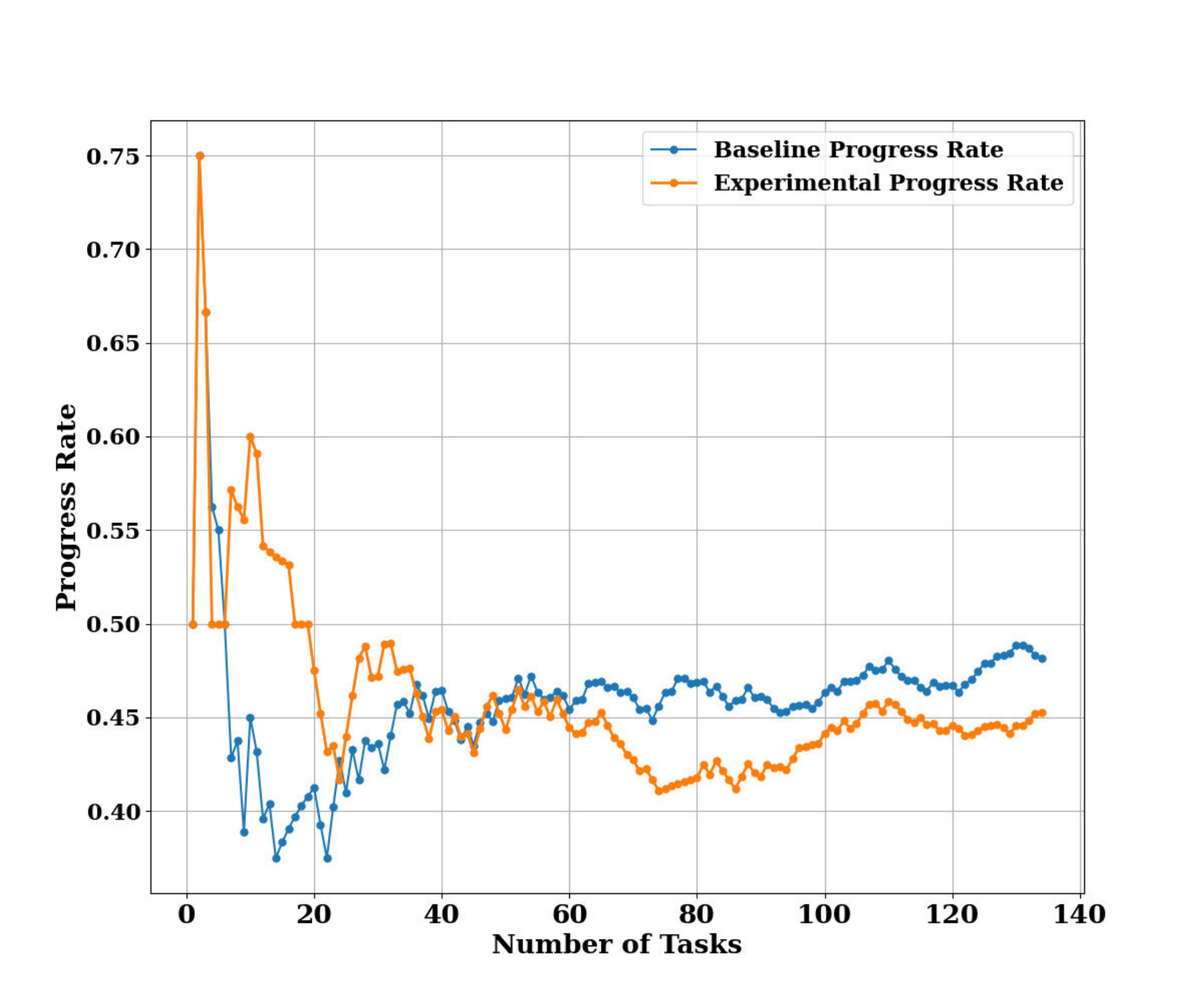}
        \caption{Progress Rate} 
        \label{fig:curve_progress}
    \end{subfigure}
    \hfill 
    \begin{subfigure}[b]{0.32\textwidth}
        \centering
        \includegraphics[width=\textwidth]{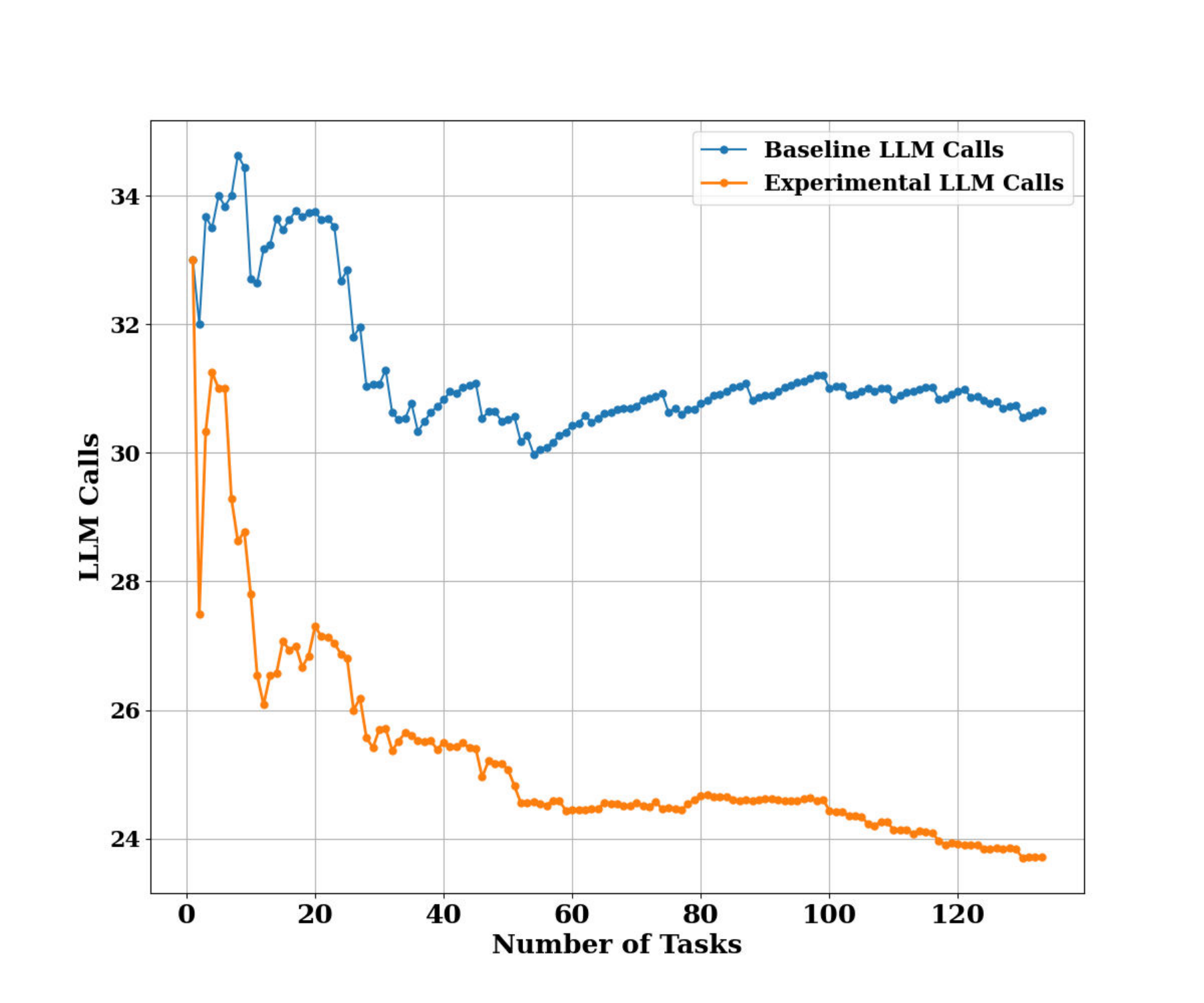}
        \caption{LLM Calls}
        \label{fig:curve_calls}
    \end{subfigure}
    \hfill
    \begin{subfigure}[b]{0.32\textwidth}
        \centering
        \includegraphics[width=\textwidth]{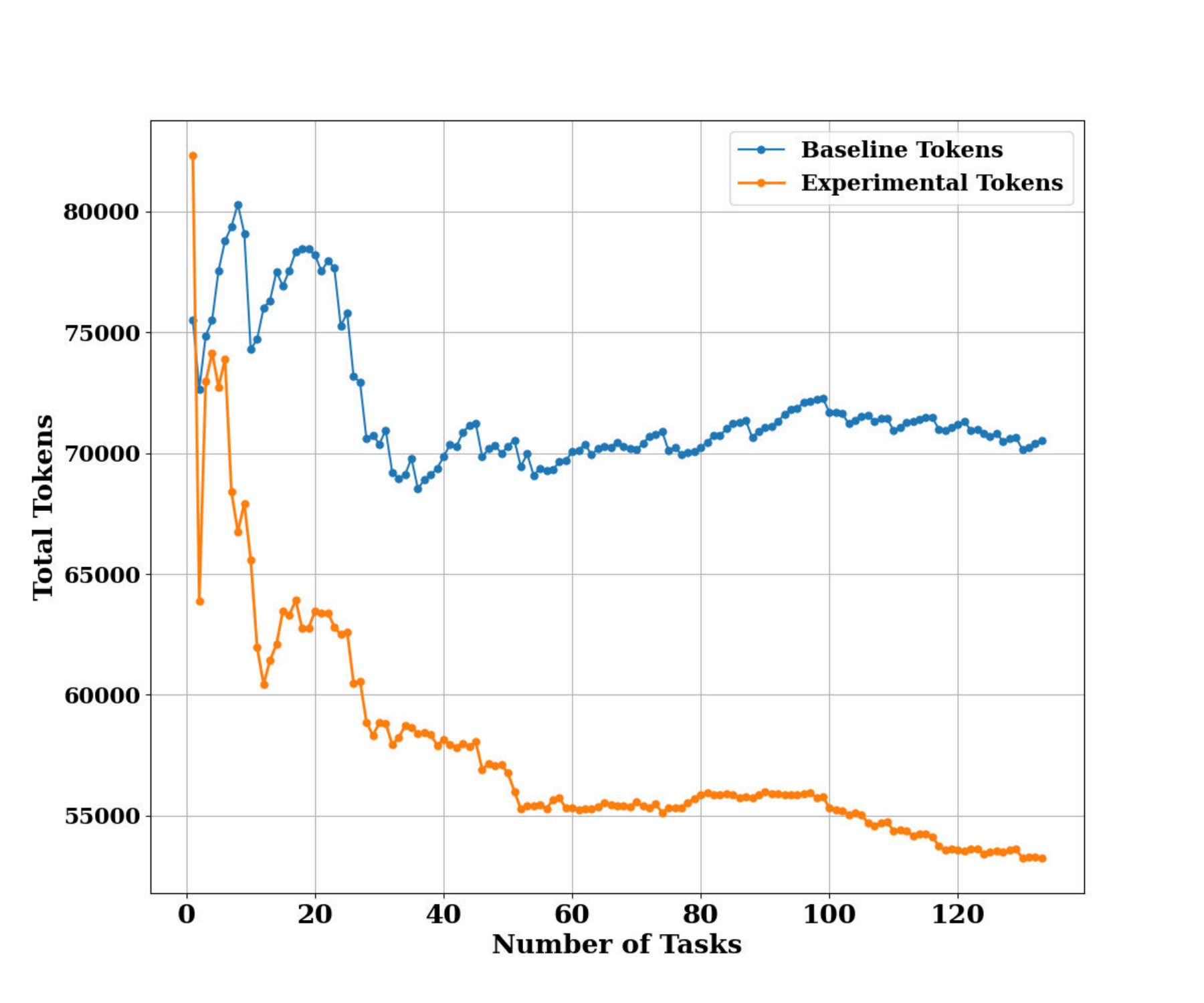}
        \caption{Token Consumption}
        \label{fig:curve_tokens}
    \end{subfigure}
    
    \caption{Dynamic learning process of Reflexion+AutoTool on the Alfworld dataset. Compared to the baseline Reflexion, our method consistently reduces (b) LLM calls and (c) token consumption, while maintaining a competitive (a) progress rate that improves over time as more trajectories are collected.}
    \label{fig:learning_curves}
\end{figure*}

To demonstrate AutoTool's dynamic learning capability and its progressive optimization effect on agent performance, we track the execution process of Reflexion+AutoTool on the Alfworld dataset, which had the most severe progress rate loss. 
Figure \ref{fig:learning_curves} shows the evolution trends of Reflexion+AutoTool and the original Reflexion in terms of Progress Rate and average LLM call counts (or Token consumption) as the number of executed tasks increases.

From the Figure \ref{fig:learning_curves}, after the experimental data converge (after 40 tasks), our method slightly underperforms the original Reflexion in terms of progress rate. However, as more execution trajectories are collected, around the 80th task execution, we observe that the performance of Reflexion+AutoTool begins to steadily improve. Considering the influence of the previous data, our method eventually maintains a comparable performance with Reflexion in the later stages of the experiment. 
More crucially, throughout the learning process, ReflexionAutoTool consistently demonstrates and gradually expands its advantage in execution efficiency. As shown in the Figure \ref{fig:learning_curves}, its average LLM call counts and Token consumption were significantly lower than those of the original Reflexion from the start, and this gap became increasingly evident as the inertia graph was refined. This highlights AutoTool's online learning capability. Even starting from scratch, it can build an effective inertia model through continuous interaction and feedback, significantly enhancing the execution efficiency of complex agents like Reflexion. This suggests that with more historical trajectory data, AutoTool's performance potential will be further unleashed.

\begin{table*}[t] 
    \centering
    \begin{tabular*}{\textwidth}{@{\extracolsep{\fill}} l l c c c c c c} 
        \toprule
        \textbf{Backbone} & \textbf{Mode} & \multicolumn{2}{c}{\textbf{AlfWorld}} & \multicolumn{2}{c}{\textbf{ScienceWorld}} & \multicolumn{2}{c}{\textbf{Academic}} \\
        \cmidrule(lr){3-4} \cmidrule(lr){5-6} \cmidrule(lr){7-8}
        & & \textbf{tried} & \textbf{success} & \textbf{tried} & \textbf{success} & \textbf{tried} & \textbf{success} \\
        \midrule
        \multirow{2}{*}{ReAct} & PDG & 68.87\% & 27.88\% & 70.22\% & 51.24\% & 74.29\% & 76.92\% \\
        & context\_filled & 31.13\% & 29.94\% & 29.78\% & 45.83\% & 25.71\% & 66.67\% \\
        \midrule
        \multirow{2}{*}{Reflexion} & PDG & 71.42\% & 22.33\% & 77.03\% & 55.90\% & 78.12\% & 76.00\% \\
        & context\_filled & 28.58\% & 25.35\% & 22.97\% & 36.46\% & 21.88\% & 71.43\% \\
        \bottomrule
    \end{tabular*}
    \caption{Parameter Filling Statistics}
    \label{tab:param_filling_acc}

\end{table*}

\section{Theoretical Analysis}
The efficiency of AutoTool is rooted in a profound insight into the agent's decision-making process, centered on the ``tool usage inertia" phenomenon. We establish the theoretical foundation for this phenomenon on principles from information theory, cognitive science, and graph search algorithms.

\textbf{Low-Entropy Markov Process of Tool Selection:}
We model the agent's tool selection sequence as a \textbf{Markov Decision Process (MDP)}. In this model, the set of available tools forms the state space $S = \{tool_1, \dots, tool_N\}$, and a sequence of invocations is a trajectory of states $(s_1, s_2, \dots, s_t)$ where $s_t \in S$. The ``tool usage inertia" observed in our work corresponds, in information theory, to a transition matrix with very low \textbf{Conditional Entropy}~\cite{shannon1948mathematical}. This is formally expressed as the uncertainty of the next state $s_{t+1}$ given the current state $s_t$:
\begin{equation} \label{eq:entropy}
H(S_{t+1}|S_t) = -\sum_{i,j \in S} p(s_i, s_j) \log_2 p(s_j|s_i)
\end{equation}
where $p(s_j|s_i)$ is the transition probability from $s_i$ to $s_j$. Our empirical findings show that for most tools $s_i$, the distribution $p(s_j|s_i)$ is highly skewed. Consequently, the system's conditional entropy is significantly lower than the maximum possible entropy for a random selection, $H_{\max} = \log_2|S|$. This condition, $H(S_{t+1}|S_t) \ll \log_2|S|$, provides the theoretical basis for why fast, non-LLM statistical prediction is feasible.

\textbf{Cognitive Heuristics:}
Priming and Habit Loops:This inertia phenomenon is analogous to concepts in cognitive science. The execution of a preceding tool acts as a \textbf{Priming Effect}. Furthermore, AutoTool's dynamic graph update mechanism mimics the \textbf{Habit Loop} (Cue $\rightarrow$ Routine $\rightarrow$ Reward)~\cite{duhigg2012power}. The feedback from the environment acts as a reward, captured by a dynamic weight update rule for the edge from tool $s_i$ to $s_j$ in our Tool Inertia Graph (TIG):
\begin{equation} \label{eq:update_rule}
w_{t+1}(s_i, s_j) = w_t(s_i, s_j) + \begin{cases} 
\Delta w_{\text{success}} & \text{if inertial call is successful} \\
-\Delta w_{\text{failure}} & \text{if inertial call fails}
\end{cases}
\end{equation}
where $\Delta w > 0$. This mechanism allows the agent to continuously learn and reinforce effective tool sequences from experience.

\textbf{Graph Search as a Decisional Shortcut:} An unconstrained LLM, at each decision step, faces a \textbf{combinatorially vast action space}. If an agent can choose from $N$ tools at each step for a sequence of length $T$, the number of potential trajectories is on the order of $N^T$. The LLM must implicitly navigate this enormous space. In contrast, by constructing historical trajectories into a Dynamic Tool Inertia Graph (TIG), AutoTool transforms this open-ended decision problem into a constrained \textbf{Graph Search} problem.

The agent's choice is guided by the Comprehensive Inertia Potential Score (CIPS), whose decision-making philosophy is akin to the evaluation function in an A* search algorithm:
\begin{equation} \label{eq:cips}
\text{CIPS}(s_{t+1}|h_t) = (1-\alpha) \cdot \text{Score}_{\text{freq}}(s_{t+1}|h_t) + \alpha \cdot \text{Score}_{\text{ctx}}(s_{t+1}|h_t)
\end{equation}
While A* aims to minimize cost ($f(n)=g(n)+h(n)$) and CIPS aims to maximize a score, their core structure combines historical information with forward-looking heuristics. Here, $\text{Score}_{\text{freq}}$ acts like the known cost $g(n)$, exploiting historical data to evaluate a path's proven reliability. Meanwhile, $\text{Score}_{\text{ctx}}$ functions as the heuristic $h(n)$, predicting future success by evaluating a tool's relevance to the current task. This A*-like mechanism allows AutoTool to balance \textbf{exploitation} (relying on 'habits') and \textbf{exploration} (responding to the current 'stimulus').

The agent then selects the tool that maximizes this score, bypassing a full LLM inference call:
\begin{equation} \label{eq:argmax}
s^* = \argmax_{s_{t+1} \in \text{Candidates}} \left( \text{CIPS}(s_{t+1}|h_t) \right)
\end{equation}
The work by Zhuang et al.~\cite{zhuang2023toolchain} also demonstrates the effectiveness of such guided search for navigating the LLM action space.

\section{Limitations}
As a data-driven acceleration method, AutoTool's performance relies on the quality and quantity of historical data, which may present cold-start challenges, particularly when dealing with extensive toolsets. Its inertia-based predictions may also be less effective in highly dynamic environments or tasks requiring intricate reasoning. A further consideration involves AutoTool’s dependency on parsing and integrating tool outputs. Designing targeted parsing functions for tools that give highly unstructured outputs may require additional engineering effort. Finally, the current manual approach to setting hyperparameters also presents an opportunity for future optimization. To advance AutoTool’s robustness and usability, future work will concentrate on exploring adaptive hyperparameter tuning and investigating deeper integration of semantic information to bolster decision-making in complex environments.

\section{Parameter Filling Accuracy}
To further validate the robustness of AutoTool, we quantitatively analyzed the accuracy of our parameter filling module. A parameter filling event is deemed successful if it leads to a tool call that is both syntactically valid and get successful environment feedback. 

Table \ref{tab:param_filling_acc} presents the proportion of attempts (tried) and success rates (success) for each parameter filling mode across different environments. We observe that the Parameter Dependency Graph (PDG) mode is predominantly utilized across all environments and backbones. Parameter filling success rates vary by environment: Academic tasks exhibit the highest success rates, likely due to their structured inputs, while AlfWorld consistently presents the lowest success rates for both PDG and context\_filled modes, suggesting inherent challenges in its open-ended and linguistically diverse environment, with ScienceWorld showing intermediate performance.

\section{Other Baseline}
Given the sequential nature of tool calls, we implemented an N-gram model (n=3) as a new baseline, which also starts collecting data from scratch. Furthermore, we conducted separate experiments for this baseline under two conditions: with and without a recovery mechanism. For parameter filling, this baseline reuses the parameter filling module from AutoTool. The final results are presented in the Table~\ref{tab:ngram_comparison}.
\begin{table}[h!] 
    \centering 
    \setlength{\tabcolsep}{3pt} 
    \begin{tabular}{
        l 
        S[table-format=1.6] 
        S[table-format=5.2] 
        S[table-format=4.2] 
        S[table-format=2.2] 
    }
        \toprule 
        {Method} & {PR} & {tok-in} & {tok-out} & {LLMC} \\
        \midrule 
        
        AutoTool        & 0.6698 & 80220 & 1324 & 18.6 \\
        Ngram\_w\_recovery           & 0.6476   & 81838 & 1403 & 18.9 \\
        Ngram\_wo\_recovery & 0.6140 & 78316 & 1135 & 18.6 \\
        
        \bottomrule
    \end{tabular}
    \caption{Comparison of AutoTool with N-gram Baselines on the ScienceWorld Dataset with Qwen2.5-32B.}
    \label{tab:ngram_comparison}
\end{table}

\section{Prompt}

\begin{figure}[htbp]
    \centering
    \includegraphics[width=\linewidth]{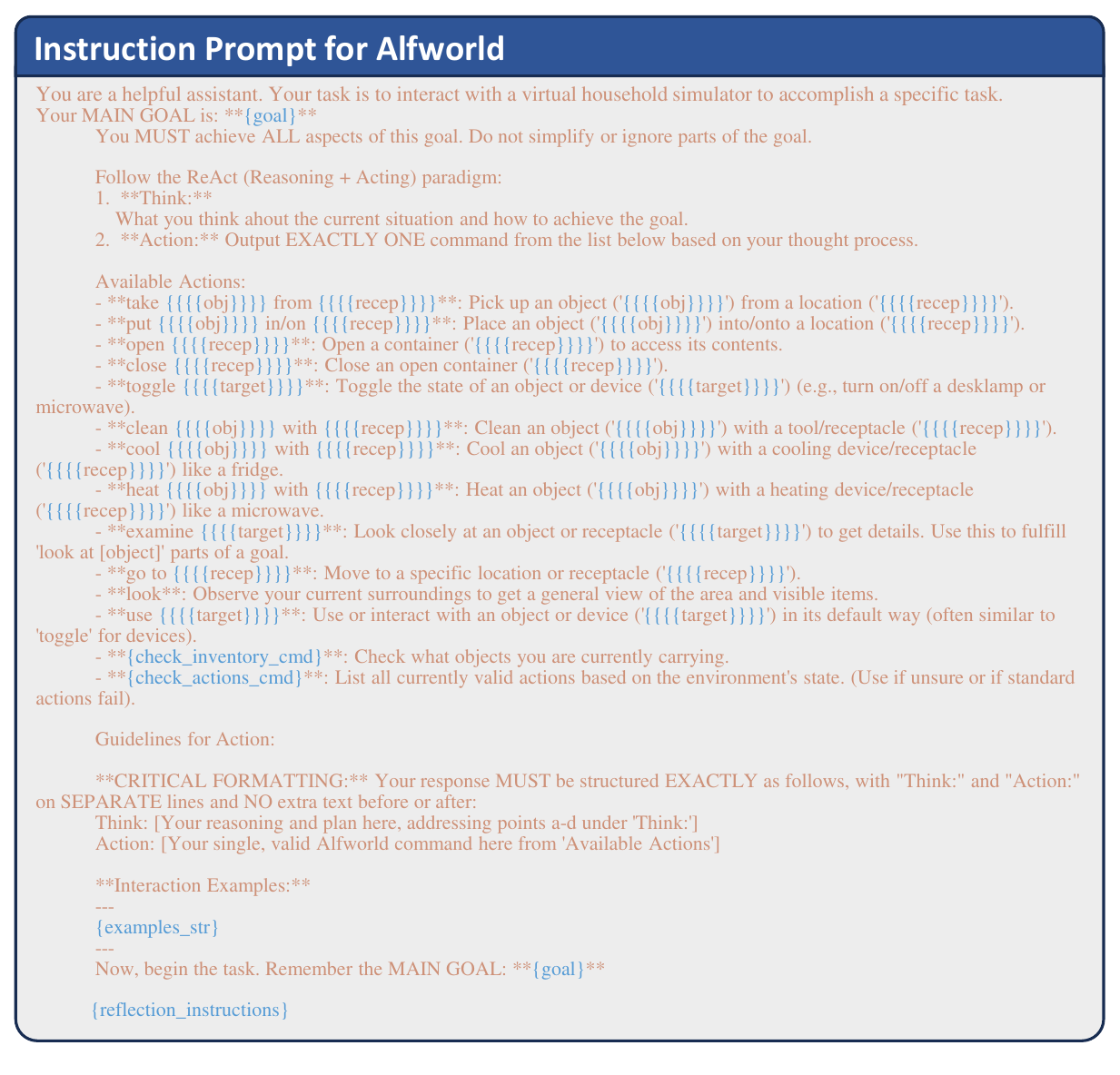} 
    \caption{Instruction prompt for alfworld}
    \label{fig:alfworld_prompt}
\end{figure}

\begin{figure}[!]
    \centering
    \includegraphics[width=\linewidth]{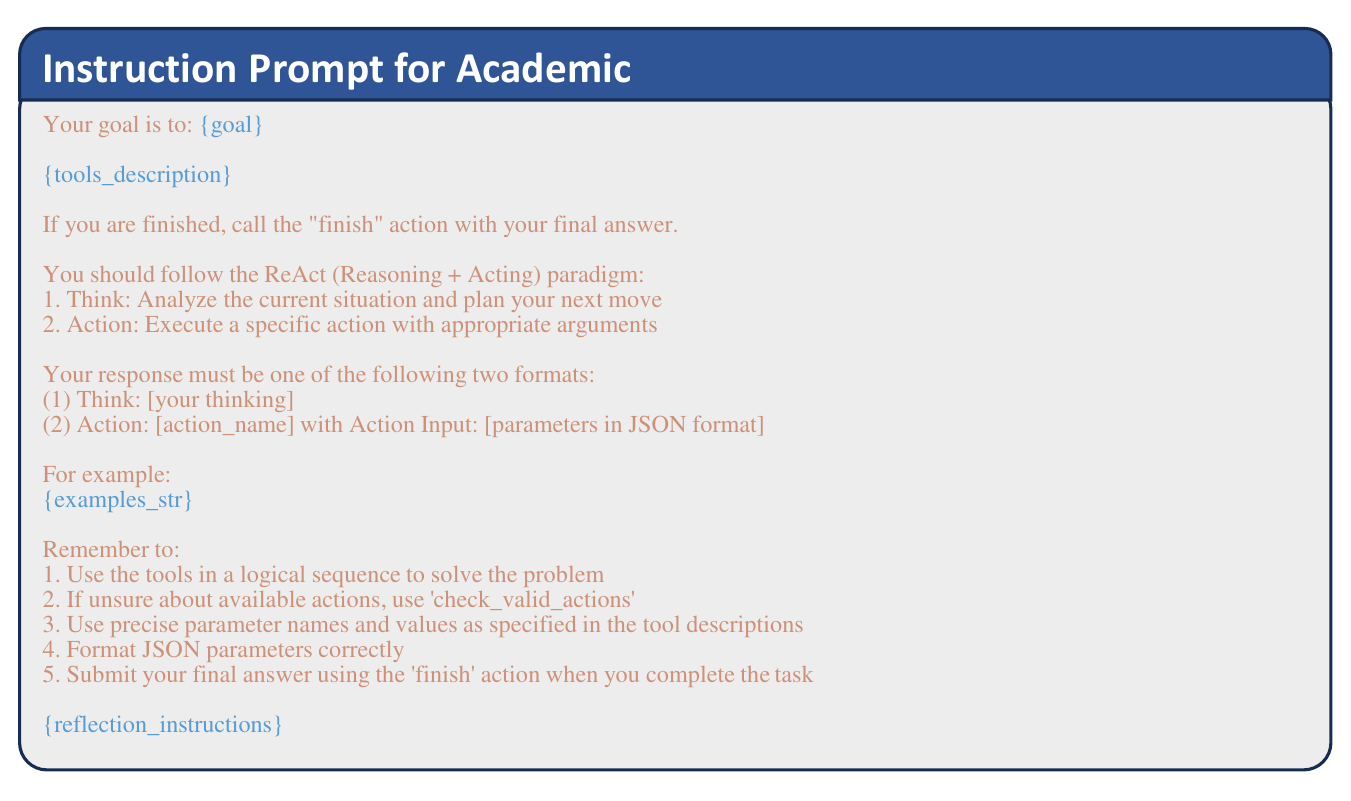}
    \caption{Instruction prompt for academic}
    \label{fig:academic_prompt}
\end{figure}

\begin{figure}[htbp]
    \centering
    \includegraphics[width=\linewidth]{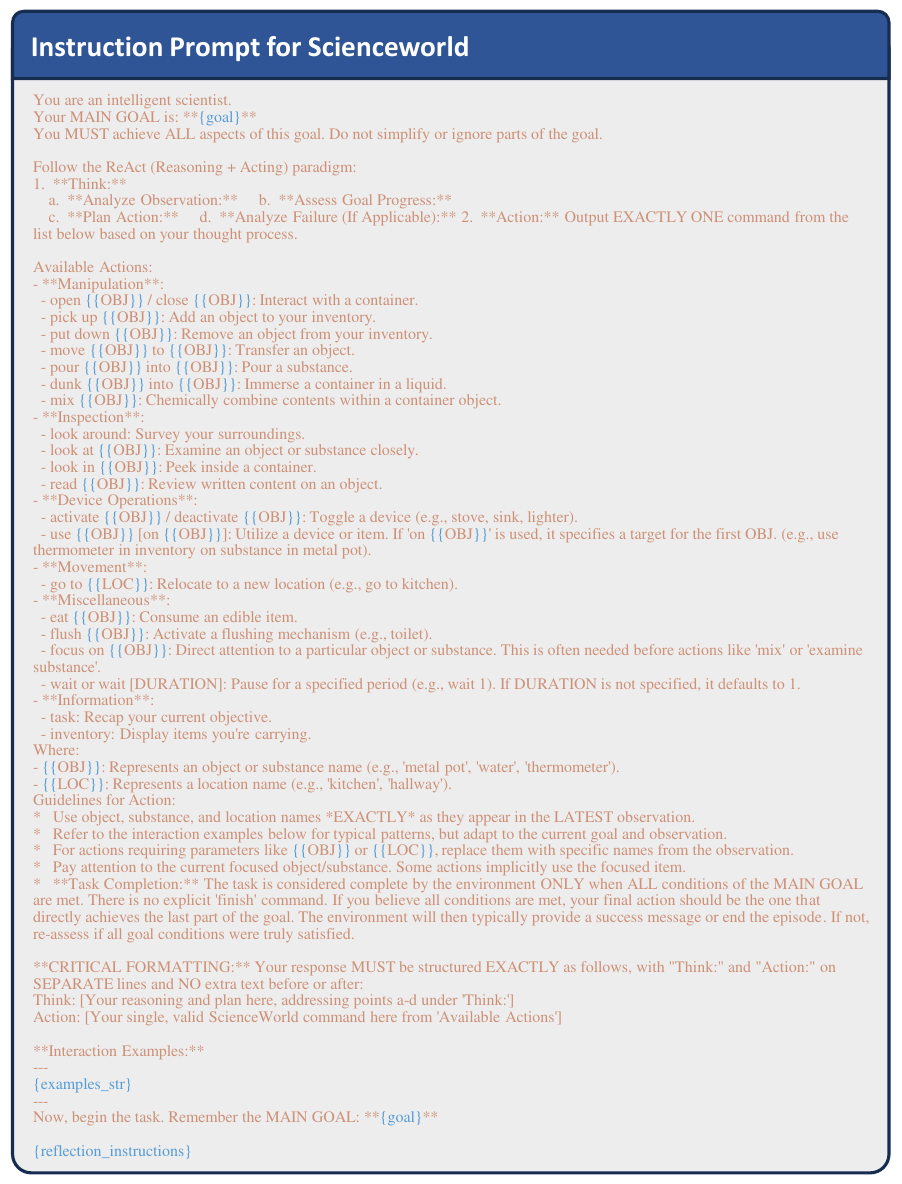} 
    \caption{Instruction prompt for scienceworld}
    \label{fig:scienceworld_prompt}
\end{figure}

We have provided the instruction prompts for the agent on three datasets. The specific details are shown in Figures ~\ref{fig:alfworld_prompt}, \ref{fig:scienceworld_prompt} ,  and \ref{fig:academic_prompt}. For the placeholders in the prompts, we can selectively fill them in. The \emph{reflection instructions} are the reflective messages generated by the LLM in Reflexion, and no reflection is provided for ReAct. Except for the Scienceworld dataset, all other datasets use the zero-shot method without providing examples.
Our AutoTool method does not modify the system prompt, that is, it uses the same instruction prompt as the baseline. 

For actions generated by inertia calls, AutoTool specifically note them when encapsulating them into temporary memory. Specifically, the prompt is:
\textbf{Think: Using graph inertia to predict next action \{predicted\_tool\} with parameters \{filled\_params\}. \textbackslash nAction: \{final\_action\}}

%% file: aaai2026.bib
@article{belcak2025small,
  title={Small Language Models are the Future of Agentic AI},
  author={Belcak, Peter and Heinrich, Greg and Diao, Shizhe and Fu, Yonggan and Dong, Xin and Muralidharan, Saurav and Lin, Yingyan Celine and Molchanov, Pavlo},
  journal={arXiv preprint arXiv:2506.02153},
  year={2025}
}

@article{team2025tongyi,
  title={Tongyi DeepResearch Technical Report},
  author={Team, Tongyi DeepResearch and Li, Baixuan and Zhang, Bo and Zhang, Dingchu and Huang, Fei and Li, Guangyu and Chen, Guoxin and Yin, Huifeng and Wu, Jialong and Zhou, Jingren and others},
  journal={arXiv preprint arXiv:2510.24701},
  year={2025}
}

@article{qian2025toolrl,
  title={Toolrl: Reward is all tool learning needs},
  author={Qian, Cheng and Acikgoz, Emre Can and He, Qi and Wang, Hongru and Chen, Xiusi and Hakkani-T{\"u}r, Dilek and Tur, Gokhan and Ji, Heng},
  journal={arXiv preprint arXiv:2504.13958},
  year={2025}
}

@article{wang2024agent,
  title={Agent workflow memory},
  author={Wang, Zora Zhiruo and Mao, Jiayuan and Fried, Daniel and Neubig, Graham},
  journal={arXiv preprint arXiv:2409.07429},
  year={2024}
}

@article{dubey2024llama,
  title={The llama 3 herd of models},
  author={Dubey, Abhimanyu and Jauhri, Abhinav and Pandey, Abhinav and Kadian, Abhishek and Al-Dahle, Ahmad and Letman, Aiesha and Mathur, Akhil and Schelten, Alan and Yang, Amy and Fan, Angela and others},
  journal={arXiv e-prints},
  pages={arXiv--2407},
  year={2024}
}

@article{yang2025qwen3,
  title={Qwen3 technical report},
  author={Yang, An and Li, Anfeng and Yang, Baosong and Zhang, Beichen and Hui, Binyuan and Zheng, Bo and Yu, Bowen and Gao, Chang and Huang, Chengen and Lv, Chenxu and others},
  journal={arXiv preprint arXiv:2505.09388},
  year={2025}
}

@article{achiam2023gpt,
  title={Gpt-4 technical report},
  author={Achiam, Josh and Adler, Steven and Agarwal, Sandhini and Ahmad, Lama and Akkaya, Ilge and Aleman, Florencia Leoni and Almeida, Diogo and Altenschmidt, Janko and Altman, Sam and Anadkat, Shyamal and others},
  journal={arXiv preprint arXiv:2303.08774},
  year={2023}
}

@inproceedings{devlin2019bert,
  title={Bert: Pre-training of deep bidirectional transformers for language understanding},
  author={Devlin, Jacob and Chang, Ming-Wei and Lee, Kenton and Toutanova, Kristina},
  booktitle={Proceedings of the 2019 conference of the North American chapter of the association for computational linguistics: human language technologies, volume 1 (long and short papers)},
  pages={4171--4186},
  year={2019}
}

@inproceedings{wang2025megaagent,
  title={MegaAgent: A large-scale autonomous LLM-based multi-agent system without predefined SOPs},
  author={Wang, Qian and Wang, Tianyu and Tang, Zhenheng and Li, Qinbin and Chen, Nuo and Liang, Jingsheng and He, Bingsheng},
  booktitle={Findings of the Association for Computational Linguistics: ACL 2025},
  pages={4998--5036},
  year={2025}
}

@article{guo2025deepseek,
  title={Deepseek-r1: Incentivizing reasoning capability in llms via reinforcement learning},
  author={Guo, Daya and Yang, Dejian and Zhang, Haowei and Song, Junxiao and Zhang, Ruoyu and Xu, Runxin and Zhu, Qihao and Ma, Shirong and Wang, Peiyi and Bi, Xiao and others},
  journal={arXiv preprint arXiv:2501.12948},
  year={2025}
}

@article{yi2024survey,
  title={A survey on recent advances in llm-based multi-turn dialogue systems},
  author={Yi, Zihao and Ouyang, Jiarui and Liu, Yuwen and Liao, Tianhao and Xu, Zhe and Shen, Ying},
  journal={arXiv preprint arXiv:2402.18013},
  year={2024}
}

@article{wei2025plangenllms,
  title={Plangenllms: A modern survey of llm planning capabilities},
  author={Wei, Hui and Zhang, Zihao and He, Shenghua and Xia, Tian and Pan, Shijia and Liu, Fei},
  journal={arXiv preprint arXiv:2502.11221},
  year={2025}
}

@article{jin2024llms,
  title={From llms to llm-based agents for software engineering: A survey of current, challenges and future},
  author={Jin, Haolin and Huang, Linghan and Cai, Haipeng and Yan, Jun and Li, Bo and Chen, Huaming},
  journal={arXiv preprint arXiv:2408.02479},
  year={2024}
}

@article{li2024personal,
  title={Personal llm agents: Insights and survey about the capability, efficiency and security},
  author={Li, Yuanchun and Wen, Hao and Wang, Weijun and Li, Xiangyu and Yuan, Yizhen and Liu, Guohong and Liu, Jiacheng and Xu, Wenxing and Wang, Xiang and Sun, Yi and others},
  journal={arXiv preprint arXiv:2401.05459},
  year={2024}
}

@article{kim2025cost,
  title={The Cost of Dynamic Reasoning: Demystifying AI Agents and Test-Time Scaling from an AI Infrastructure Perspective},
  author={Kim, Jiin and Shin, Byeongjun and Chung, Jinha and Rhu, Minsoo},
  journal={arXiv preprint arXiv:2506.04301},
  year={2025}
}

@inproceedings{yao2023react,
  title={React: Synergizing reasoning and acting in language models},
  author={Yao, Shunyu and Zhao, Jeffrey and Yu, Dian and Du, Nan and Shafran, Izhak and Narasimhan, Karthik and Cao, Yuan},
  booktitle={International Conference on Learning Representations (ICLR)},
  year={2023}
}

@article{paranjape2023art,
  title={Art: Automatic multi-step reasoning and tool-use for large language models},
  author={Paranjape, Bhargavi and Lundberg, Scott and Singh, Sameer and Hajishirzi, Hannaneh and Zettlemoyer, Luke and Ribeiro, Marco Tulio},
  journal={arXiv preprint arXiv:2303.09014},
  year={2023}
}

@article{qin2024tool,
  title={Tool learning with foundation models},
  author={Qin, Yujia and Hu, Shengding and Lin, Yankai and Chen, Weize and Ding, Ning and Cui, Ganqu and Zeng, Zheni and Zhou, Xuanhe and Huang, Yufei and Xiao, Chaojun and others},
  journal={ACM Computing Surveys},
  volume={57},
  number={4},
  pages={1--40},
  year={2024},
  publisher={ACM New York, NY}
}

@article{qu2025tool,
  title={Tool learning with large language models: A survey},
  author={Qu, Changle and Dai, Sunhao and Wei, Xiaochi and Cai, Hengyi and Wang, Shuaiqiang and Yin, Dawei and Xu, Jun and Wen, Ji-Rong},
  journal={Frontiers of Computer Science},
  volume={19},
  number={8},
  pages={198343},
  year={2025},
  publisher={Springer}
}

@article{shen2023hugginggpt,
  title={Hugginggpt: Solving ai tasks with chatgpt and its friends in hugging face},
  author={Shen, Yongliang and Song, Kaitao and Tan, Xu and Li, Dongsheng and Lu, Weiming and Zhuang, Yueting},
  journal={Advances in Neural Information Processing Systems},
  volume={36},
  pages={38154--38180},
  year={2023}
}

@article{song2023restgpt,
  title={Restgpt: Connecting large language models with real-world restful apis},
  author={Song, Yifan and Xiong, Weimin and Zhu, Dawei and Wu, Wenhao and Qian, Han and Song, Mingbo and Huang, Hailiang and Li, Cheng and Wang, Ke and Yao, Rong and others},
  journal={arXiv preprint arXiv:2306.06624},
  year={2023}
}

@inproceedings{zhuge2024gptswarm,
  title={Gptswarm: Language agents as optimizable graphs},
  author={Zhuge, Mingchen and Wang, Wenyi and Kirsch, Louis and Faccio, Francesco and Khizbullin, Dmitrii and Schmidhuber, J{\"u}rgen},
  booktitle={Forty-first International Conference on Machine Learning},
  year={2024}
}

@article{zhuang2023toolchain,
  title={Toolchain*: Efficient action space navigation in large language models with a* search},
  author={Zhuang, Yuchen and Chen, Xiang and Yu, Tong and Mitra, Saayan and Bursztyn, Victor and Rossi, Ryan A and Sarkhel, Somdeb and Zhang, Chao},
  journal={arXiv preprint arXiv:2310.13227},
  year={2023}
}

@article{yao2023tree,
  title={Tree of thoughts: Deliberate problem solving with large language models},
  author={Yao, Shunyu and Yu, Dian and Zhao, Jeffrey and Shafran, Izhak and Griffiths, Tom and Cao, Yuan and Narasimhan, Karthik},
  journal={Advances in neural information processing systems},
  volume={36},
  pages={11809--11822},
  year={2023}
}

@article{schick2023toolformer,
  title={Toolformer: Language models can teach themselves to use tools},
  author={Schick, Timo and Dwivedi-Yu, Jane and Dess{\`\i}, Roberto and Raileanu, Roberta and Lomeli, Maria and Hambro, Eric and Zettlemoyer, Luke and Cancedda, Nicola and Scialom, Thomas},
  journal={Advances in Neural Information Processing Systems},
  volume={36},
  pages={68539--68551},
  year={2023}
}

@article{patil2024gorilla,
  title={Gorilla: Large language model connected with massive apis},
  author={Patil, Shishir G and Zhang, Tianjun and Wang, Xin and Gonzalez, Joseph E},
  journal={Advances in Neural Information Processing Systems},
  volume={37},
  pages={126544--126565},
  year={2024}
}

@article{qin2023toolllm,
  title={Toolllm: Facilitating large language models to master 16000+ real-world apis},
  author={Qin, Yujia and Liang, Shihao and Ye, Yining and Zhu, Kunlun and Yan, Lan and Lu, Yaxi and Lin, Yankai and Cong, Xin and Tang, Xiangru and Qian, Bill and others},
  journal={arXiv preprint arXiv:2307.16789},
  year={2023}
}

@article{koh2024tree,
  title={Tree search for language model agents},
  author={Koh, Jing Yu and McAleer, Stephen and Fried, Daniel and Salakhutdinov, Ruslan},
  journal={arXiv preprint arXiv:2407.01476},
  year={2024}
}

@article{du2024anytool,
  title={Anytool: Self-reflective, hierarchical agents for large-scale api calls},
  author={Du, Yu and Wei, Fangyun and Zhang, Hongyang},
  journal={arXiv preprint arXiv:2402.04253},
  year={2024}
}

@article{liu2024tool,
  title={Tool-Planner: Task Planning with Clusters across Multiple Tools},
  author={Liu, Yanming and Peng, Xinyue and Cao, Jiannan and Bo, Shi and Zhang, Yuwei and Zhang, Xuhong and Cheng, Sheng and Wang, Xun and Yin, Jianwei and Du, Tianyu},
  journal={arXiv preprint arXiv:2406.03807},
  year={2024}
}

@article{shinn2023reflexion,
  title={Reflexion: Language agents with verbal reinforcement learning},
  author={Shinn, Noah and Cassano, Federico and Gopinath, Ashwin and Narasimhan, Karthik and Yao, Shunyu},
  journal={Advances in Neural Information Processing Systems},
  volume={36},
  pages={8634--8652},
  year={2023}
}

@article{ma2024agentboard,
  title={Agentboard: An analytical evaluation board of multi-turn llm agents},
  author={Ma, Chang and Zhang, Junlei and Zhu, Zhihao and Yang, Cheng and Yang, Yujiu and Jin, Yaohui and Lan, Zhenzhong and Kong, Lingpeng and He, Junxian},
  journal={arXiv preprint arXiv:2401.13178},
  year={2024}
}

@article{shridhar2020alfworld,
  title={Alfworld: Aligning text and embodied environments for interactive learning},
  author={Shridhar, Mohit and Yuan, Xingdi and C{\^o}t{\'e}, Marc-Alexandre and Bisk, Yonatan and Trischler, Adam and Hausknecht, Matthew},
  journal={arXiv preprint arXiv:2010.03768},
  year={2020}
}

@article{wang2022scienceworld,
  title={Scienceworld: Is your agent smarter than a 5th grader?},
  author={Wang, Ruoyao and Jansen, Peter and C{\^o}t{\'e}, Marc-Alexandre and Ammanabrolu, Prithviraj},
  journal={arXiv preprint arXiv:2203.07540},
  year={2022}
}

@misc{langchain2023,
  author = {LangChain},
  title = {Langchain: Build context-aware reasoning applications},
  year = {2023},
  url = {https://github.com/langchain-ai/langchain},
  note = {URL: \url{https://github.com/langchain-ai/langchain}}
}

@article{hong2023metagpt,
  title={Metagpt: Meta programming for multi-agent collaborative framework},
  author={Hong, Sirui and Zheng, Xiawu and Chen, Jonathan and Cheng, Yuheng and Wang, Jinlin and Zhang, Ceyao and Wang, Zili and Yau, Steven Ka Shing and Lin, Zijuan and Zhou, Liyang and others},
  journal={arXiv preprint arXiv:2308.00352},
  volume={3},
  number={4},
  pages={6},
  year={2023}
}

@article{gao2021simcse,
  title={Simcse: Simple contrastive learning of sentence embeddings},
  author={Gao, Tianyu and Yao, Xingcheng and Chen, Danqi},
  journal={arXiv preprint arXiv:2104.08821},
  year={2021}
}

@inproceedings{kim2024llm,
  title={An llm compiler for parallel function calling},
  author={Kim, Sehoon and Moon, Suhong and Tabrizi, Ryan and Lee, Nicholas and Mahoney, Michael W and Keutzer, Kurt and Gholami, Amir},
  booktitle={Forty-first International Conference on Machine Learning},
  year={2024}
}

@article{liu2024toolnet,
  title={Toolnet: Connecting large language models with massive tools via tool graph},
  author={Liu, Xukun and Peng, Zhiyuan and Yi, Xiaoyuan and Xie, Xing and Xiang, Lirong and Liu, Yuchen and Xu, Dongkuan},
  journal={arXiv preprint arXiv:2403.00839},
  year={2024}
}

@article{wu2024can,
  title={Can graph learning improve planning in LLM-based agents?},
  author={Wu, Xixi and Shen, Yifei and Shan, Caihua and Song, Kaitao and Wang, Siwei and Zhang, Bohang and Feng, Jiarui and Cheng, Hong and Chen, Wei and Xiong, Yun and others},
  journal={Advances in Neural Information Processing Systems},
  volume={37},
  pages={5338--5383},
  year={2024}
}

@article{shannon1948mathematical,
  title={A mathematical theory of communication},
  author={Shannon, Claude E},
  journal={The Bell system technical journal},
  volume={27},
  number={3},
  pages={379--423},
  year={1948},
  publisher={Nokia Bell Labs}
}

@book{duhigg2012power,
  title={The power of habit: Why we do what we do in life and business},
  author={Duhigg, Charles},
  volume={34},
  number={10},
  year={2012},
  publisher={Random House}
}

@article{hui2024qwen2,
  title={Qwen2. 5-coder technical report},
  author={Hui, Binyuan and Yang, Jian and Cui, Zeyu and Yang, Jiaxi and Liu, Dayiheng and Zhang, Lei and Liu, Tianyu and Zhang, Jiajun and Yu, Bowen and Lu, Keming and others},
  journal={arXiv preprint arXiv:2409.12186},
  year={2024}
}

@article{liu2024deepseek,
  title={Deepseek-v3 technical report},
  author={Liu, Aixin and Feng, Bei and Xue, Bing and Wang, Bingxuan and Wu, Bochao and Lu, Chengda and Zhao, Chenggang and Deng, Chengqi and Zhang, Chenyu and Ruan, Chong and others},
  journal={arXiv preprint arXiv:2412.19437},
  year={2024}
}

@misc{dubey2024llama3,
      title={The Llama 3 Herd of Models}, 
      author={Abhimanyu Dubey and Abhinav Jauhri and Abhinav Pandey and Abhishek Kadian and Ahmad Al-Dahle and Aiesha Letman and Akhil Mathur and Alan Schelten and Amy Yang and Angela Fan and et al.},
      year={2024},
      eprint={2407.21783},
      archivePrefix={arXiv},
      primaryClass={cs.CL}
}
